\documentclass[10pt,journal,compsoc]{IEEEtran}

\usepackage{cite}
\usepackage{amsmath,amssymb}
\usepackage{graphicx}
\usepackage{tikz}
\usepackage{booktabs}
\usepackage{multirow}
\usepackage{xcolor}
\usepackage{capt-of}
\usepackage{orcidlink}
\hypersetup{hidelinks}

\newcommand{\corrauth}{\textsuperscript{\textdagger}}
\newcommand{\projectlead}{\textsuperscript{\textdaggerdbl}}

\usepackage{etoolbox}
\makeatletter
\patchcmd{\section}{-3.5ex plus -2ex minus -1.5ex}{-1.5ex plus -.5ex minus -.3ex}{}{}
\patchcmd{\subsection}{-3.5ex plus -1.5ex minus -1.5ex}{-1.5ex plus -.5ex minus -.3ex}{}{}
\patchcmd{\section}{0.7ex plus 1ex minus 0ex}{0.3ex plus .2ex minus 0ex}{}{}
\patchcmd{\subsection}{0.7ex plus .5ex minus 0ex}{0.3ex plus .2ex minus 0ex}{}{}
\makeatother

\begin{document}

\title{\texorpdfstring{\raisebox{-0.32ex}{\includegraphics[width=0.35\textwidth,trim=65bp 160bp 55bp 180bp,clip]{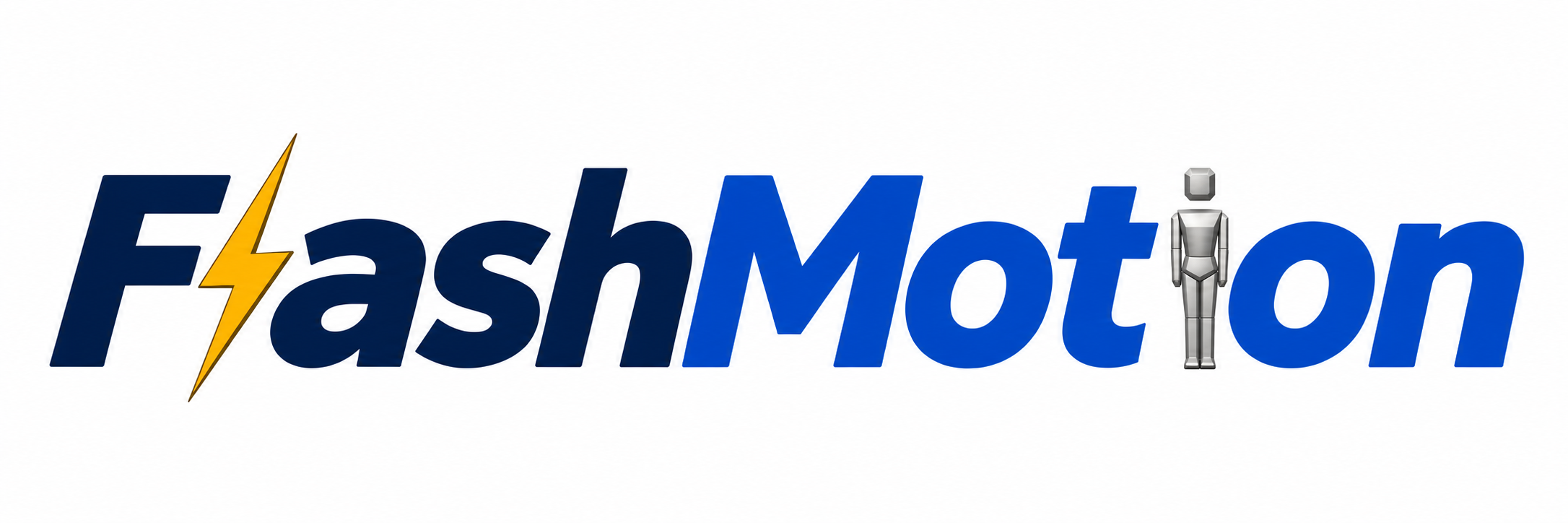}}}{FlashMotion}: \\
Ultra-Fast Long-Term 3D Motion Generation}

\author{\small Kaixing Yang\orcidlink{0000-0002-3879-7225},
Xulong Tang\projectlead\orcidlink{0009-0005-0857-1056}, Ronghui Li\projectlead\orcidlink{0000-0001-7703-9315}, Xiangyue Zhang\orcidlink{0009-0002-5642-9474},
Ziqiao Peng\orcidlink{0009-0000-1341-9618},
Mingshuang Luo,
Chenxi Zhao\orcidlink{0009-0005-9898-8049}, Haoyu Wu\orcidlink{0009-0003-0618-5683},
	\\ Ju Fan\orcidlink{0000-0003-4729-9903},
Hongyan Liu\corrauth\orcidlink{0000-0002-4902-1078},
Puwei Wang\corrauth\orcidlink{0000-0002-2949-5380},
Zijiao Zeng\corrauth\orcidlink{0009-0008-0378-373X},
and Jun He\orcidlink{0000-0003-1511-7554}\thanks{\corrauth Corresponding authors: Hongyan Liu, Puwei Wang, and Zijiao Zeng.}\thanks{\projectlead Project leaders: Xulong Tang and Ronghui Li.}\thanks{This work was conducted at Tencent IEG Motus Group.}\thanks{Kaixing Yang, Ziqiao Peng, Ju Fan, Puwei Wang, and Jun He are with Renmin University of China, Beijing, China (e-mail: \{yangkaixing,pengziqiao,fanj,wangpuwei,hejun\}@ruc.edu.cn). Xulong Tang is with The University of Texas at Dallas (e-mail: Xulong.Tang@UTDallas.edu). Ronghui Li and Hongyan Liu are with Tsinghua University, Beijing, China (e-mail: lrh22@mails.tsinghua.edu.cn; liuhy@sem.tsinghua.edu.cn). Xiangyue Zhang is with Wuhan University, Wuhan, China (e-mail: xiangyuezhang@whu.edu.cn). Mingshuang Luo and Zijiao Zeng are with Tencent, China (e-mail: luomingshuang841@gmail.com; \mbox{zijiaozeng@tencent.com}). Chenxi Zhao is with Nankai University, Tianjin, China (e-mail: zhaochenxi515@126.com). Haoyu Wu is with The University of Hong Kong, Hong Kong, China (e-mail: wuhaoyu556@connect.hku.hk).}\thanks{This work was supported in part by the National Natural Science Foundation of China under Grants 62436010, 72572090, 62572474, and 62172421.}}

\markboth{IEEE Transactions on Pattern Analysis and Machine Intelligence}{Yang \MakeLowercase{\textit{et al.}}: \textbf{FlashMotion}: Ultra-Fast Long-Term 3D Motion Generation}

\IEEEaftertitletext{\vspace{-28pt}}
\maketitle
\raggedbottom
\setlength{\abovedisplayskip}{4.5pt plus 1pt minus 1pt}
\setlength{\belowdisplayskip}{4.5pt plus 1pt minus 1pt}
\setlength{\abovedisplayshortskip}{3pt plus 1pt minus 1pt}
\setlength{\belowdisplayshortskip}{4pt plus 1pt minus 1pt}

\begin{abstract}
3D human motion generation plays a crucial role in game development, extended reality (XR), digital humans, and embodied simulation.
Beyond motion generation, practical pipelines include rigging, retargeting, and rendering, making ultra-fast long-term 3D motion generation without compromising quality valuable for real-time interactive applications.
Despite promising progress, high-quality long-horizon motion generation under tight inference budgets remains challenging, owing to 2 coupled bottlenecks: (1) generative strategies that require many sampling steps for high-quality 3D motion generation and (2) model architectures with substantial per-step computational and memory costs for long sequences.
We present \textbf{FlashMotion}, a co-designed
framework for high-quality long-horizon 3D motion generation under a strict inference budget, covering text-to-motion, music-to-dance, and speech-to-gesture tasks.
To enhance \textbf{trajectory-level efficiency}, we introduce a 3D motion-space MeanFlow formulation with kinematic and dynamic consistency constraints for few-step generation, together with an Adaptive Soft masking mechanism for temporally coherent motion editing.
To improve \textbf{per-step efficiency}, we design a lightweight non-autoregressive BiMamba architecture with condition-guided state selection and AdaIN-based condition fusion mechanism, enhancing cross-modal alignment and generalization to longer motion sequences.
Extensive experiments demonstrate that \textbf{FlashMotion} achieves overall state-of-the-art performance in both quality and efficiency across text-to-motion, speech-to-gesture, and music-to-dance generation.
The project page is available at \href{https://sun-happy-ykx.github.io/FlashMotion-Project/}{\textcolor{blue}{\textbf{FlashMotion}}}.
\end{abstract}


\input{teaser_comic_vector.tex}

\section{Introduction}
\IEEEPARstart{W}{ith} advances in AI-generated content and cross-modal motion modeling~\cite{yang2026omnidance,yang2024beatdance}, 3D human motion generation has attracted increasing interest for translating text, speech, and music into expressive human motion. 
This capability supports applications in robot motion generation, extended reality (XR) character animation, and mobile motion interaction (Fig.~\ref{fig:applications}).
These applications require responsive and sustained interaction. As a key enabler of such interaction, a broadly applicable 3D motion generation method should satisfy 3 key requirements:
 \textbf{(1) Motion Quality}: generating physically plausible and semantically expressive motion that aligns with the input conditions; \textbf{(2) Generation Efficiency}: achieving low inference latency and memory consumption to leave sufficient computational resources for other pipeline stages, including rigging, retargeting, and rendering; and \textbf{(3) Long-Horizon Generation}: sustaining high motion fidelity and smooth temporal transitions over extended sequences.

Across text-to-motion, speech-to-gesture, and music-to-dance tasks, 3D human motion generation has advanced along 2 complementary dimensions.
\textbf{(1) Generative Strategy.} Representative strategies include direct motion regression~\cite{ahuja2019language2pose,li2021ai}, adversarial learning~\cite{yang2025cohedancers,yoon2020speech}, autoregressive generation~\cite{zhang2023generating,yang2025megadance}, diffusion~\cite{tevet2023human,zhu2023taming}, and flow matching~\cite{canalescuba2025flowmotion,yang2025flowerdance}, with different trade-offs between generation quality and sampling efficiency.
\textbf{(2) Model Architecture.} Temporal backbones range from recurrent networks, including LSTMs~\cite{fragkiadaki2015recurrent,martinez2017human}, to Transformers~\cite{zhang2023generating,li2021ai} and Mamba-based models~\cite{xu2024mambatalk,yang2026tokendance}, with different capabilities for temporal modeling and conditional integration.
However, high-quality long-horizon generation under a strict inference budget remains challenging due to 2 coupled bottlenecks: (1) generative strategies that require many sampling steps for high-quality 3D motion generation and (2) model architectures with substantial per-step computational and memory costs for long sequences.
These costs leave limited computational resources for other production stages, such as rigging, retargeting, and rendering, potentially degrading the user experience in interactive applications.

To this end, we present \textbf{FlashMotion}, a co-designed framework for high-quality long-horizon 3D motion generation under a strict inference budget. The framework supports diverse conditional motion generation tasks, including text-to-motion, music-to-dance, and speech-to-gesture.
Our key observation is that end-to-end generation efficiency is jointly determined by (1) the number of sampling steps and (2) the computational cost of each step.
Optimizing either factor alone limits practical speedups: few-step sampling remains costly with an expensive backbone, while a lightweight architecture still incurs substantial latency when many sampling steps are required.
\textbf{FlashMotion} therefore jointly optimizes trajectory-level and per-step efficiency.
To improve trajectory-level efficiency, we introduce a 3D motion-space MeanFlow formulation with kinematic and dynamic consistency constraints for few-step generation, along with an Adaptive Soft masking mechanism for temporally coherent motion editing. To improve per-step efficiency, we design a lightweight non-autoregressive architecture that combines a BiMamba-based backbone with condition-guided state selection and AdaIN-based condition fusion.
Together, \textbf{FlashMotion} achieves state-of-the-art motion quality across all 3 tasks, generating physically plausible, semantically expressive motion at leading throughput with a compact $\sim$60M-parameter model. Fig.~\ref{fig: teaser} presents quality--throughput comparisons with representative baselines (top) and conditioned motion examples (bottom). Panels (a)-(c) report results on BABEL, BEAT2, and FineDance, respectively. FPS is measured for 10,000-frame outputs.

\textbf{Trajectory-level efficiency via generative strategy.}
To reduce sampling steps without sacrificing motion quality, \textbf{FlashMotion} combines a 3D motion-space MeanFlow formulation with kinematic and dynamic consistency constraints.
From a kinematic perspective, coherent 3D human motion requires coordinated joint trajectories and smooth temporal evolution, motivating a generative process that continuously transforms noise into structured motion. Flow Matching provides a natural framework for modeling this transformation through a learned velocity field~\cite{lipman2022flow,liu2022flow}. However, standard Flow Matching learns instantaneous velocities, whereas few-step sampling requires accurate finite-interval updates, making generation sensitive to numerical integration errors.
MeanFlow~\cite{geng2025mean} addresses this limitation by modeling interval-averaged velocity. Nevertheless, directly predicting velocity makes it less straightforward to impose structural constraints on the resulting 3D motion. Since 3D human motion representation is subject to tightly coupled kinematic and dynamic constraints, errors in large-step updates can manifest as temporal jitter, root drift, foot sliding, and implausible articulations. This motivates predicting clean motion explicitly, where motion-specific supervision can be applied directly, while deriving the velocity required for sampling from the prediction.
We therefore introduce a 3D motion-space MeanFlow formulation that takes the noisy motion $z_t$ and the time pair $(t,r)$ as inputs and predicts clean motion $\hat{z}_0$. During training, we construct the corresponding intermediate state $\hat{z}_r$ by interpolating between $z_t$ and $\hat{z}_0$, and use its displacement from $z_t$ to parameterize the interval-averaged velocity for MeanFlow training. This formulation also enables direct supervision in 3D motion space: reconstruction and forward-kinematics losses encourage faithful motion recovery and kinematic consistency, while velocity and dynamics losses regularize temporal evolution. At inference, the predicted clean motion is converted into the corresponding interval-averaged velocity for few-step ODE sampling.
In addition, \textbf{FlashMotion} provides flexible motion editing to support human-in-the-loop authoring and responsive streaming applications. Specifically, we apply an Adaptive Soft mask during sampling, which gradually relaxes the constraints over time and alleviates the boundary discontinuities caused by hard masks in the few-step setting.

\textbf{Per-step efficiency via model architecture.}
We design a lightweight non-autoregressive BiMamba architecture with condition-guided state selection and AdaIN-based condition fusion.
BiMamba is well suited to temporally coherent 3D human motion generation because its sequential inductive bias better captures local temporal dependencies than attention-only backbones, while retaining linear-time complexity $\mathcal{O}(n)$ compared with the $\mathcal{O}(n^2)$ cost of Transformers~\cite{vaswani2017attention}. Its bidirectional design further improves temporal coherence and condition-motion alignment, and the non-autoregressive formulation avoids the latency of sequential decoding while mitigating exposure bias in autoregressive-based methods~\cite{siyao2022bailando,yang2025megadance} and boundary artifacts in iterative inpainting-based methods~\cite{tseng2023edge,liu2025gcdance}.  
Condition-guided state selection regulates temporal information flow, while AdaIN modulates motion features through frame-wise scaling and shifting. Shared projections support variable sequence lengths without constructing pairwise cross-attention matrices.
Together, these architectural choices reduce per-step computational and memory costs while supporting high-quality long-horizon motion generation.

This work extends our previous work, FlowerDance~\cite{yang2025flowerdance}, in 4 aspects: (1) \textbf{Task Coverage}: generalizing the method to conditional 3D human motion generation, including text-to-motion, speech-to-gesture, and music-to-dance; (2) \textbf{Generative Strategy}: reformulating MeanFlow in 3D motion space through explicit clean-motion prediction; (3) \textbf{Model Architecture}: introducing condition-guided state selection and AdaIN-based condition fusion to enhance cross-modal alignment and long-sequence generalization; and (4) \textbf{Experiments}: expanding evaluations across tasks and targeted ablation studies, alongside application demonstrations.

In summary, our main contributions are as follows:
\begin{itemize}[\setlength{\topsep}{2pt}]
    \item We present \textbf{FlashMotion}, a co-designed framework for high-quality long-horizon 3D motion generation under a strict inference budget. By jointly optimizing trajectory-level sampling efficiency and per-step inference efficiency, \textbf{FlashMotion} achieves state-of-the-art (SOTA) performance in both quality and efficiency across text-to-motion, speech-to-gesture, and music-to-dance generation.

    \item We propose a few-step 3D motion generation strategy comprising: (1) a 3D motion-space MeanFlow formulation with kinematic and dynamic consistency constraints, and (2) Adaptive Soft masking for coherent motion editing.

    \item We design a lightweight non-autoregressive architecture for efficient long-horizon motion generation, comprising: (1) a BiMamba-based backbone, (2) a condition-guided state selection mechanism, and (3) AdaIN-based cross-modal alignment.
\end{itemize}

\section{Related Work}
\subsection{Flow Matching Generative Models}
Flow Matching (FM)~\cite{lipman2022flow} learns continuous transport from noise to data, with substantial methodological advances~\cite{liu2022flow,geng2025mean} and broad generative applications~\cite{esser2024scaling,le2023voicebox}, while its adoption in 3D human motion generation remains nascent~\cite{hu2023motion,mehta2024unified,yang2025flowerdance}.

\textbf{Methodological Advances.}
FM enables simulation-free training through velocity regression~\cite{lipman2022flow}. Rectified Flow~\cite{liu2022flow} promotes straighter transport trajectories through reflow, while optimal-transport conditional flow matching (OT-CFM)~\cite{tong2024improving} uses minibatch transport couplings to simplify learned flows. To reduce sampling cost, Consistency Flow Matching~\cite{yang2024consistency} encourages velocity consistency, Flow Map Matching~\cite{boffi2025flow} learns mappings between pairs of generation times, and Shortcut Models~\cite{frans2025one} condition predictions on the desired step size. Complementing these approaches, GMFlow~\cite{chen2025gaussian} models multimodal velocity distributions with Gaussian mixtures. MeanFlow~\cite{geng2025mean} instead learns interval-averaged velocities, directly targeting finite-interval updates for 1- or few-step generation.

\textbf{Applications Across Domains.}
FM has demonstrated its scalability across diverse generative tasks. In image generation, Stable Diffusion 3~\cite{esser2024scaling} combines rectified flow with large-scale multimodal Transformer training. Video applications include sparsely conditioned video prediction~\cite{davtyan2023efficient}, audio-driven portrait animation with FLOAT~\cite{ki2024float}, and multiscale video synthesis with Pyramid Flow Matching~\cite{jin2024pyramidal}. In audio, Voicebox~\cite{le2023voicebox} and Audiobox~\cite{vyas2023audiobox} employ FM for speech and audio generation, while Matcha-TTS~\cite{mehta2024matcha} and VoiceFlow~\cite{guo2024voiceflow} explore efficient text-to-speech synthesis.

Few studies have explored FM for 3D human motion synthesis, yet ultra-few-step long-horizon generation remains challenging: (1) approximately 50 steps for FlowMotion~\cite{canalescuba2025flowmotion}, Match-TTSG~\cite{mehta2024unified}, and UMF's prior generation stage~\cite{huang2026unified}, and (2) 10 steps for Motion Flow Matching~\cite{hu2023motion}, FloodDiffusion~\cite{cai2026flooddiffusion}, and FlowerDance~\cite{yang2025flowerdance}. 

\begin{figure*}[t]
    \centering    \resizebox{0.90\linewidth}{!}{\begin{tikzpicture}[x=1bp,y=-1bp]
\node[anchor=north west,inner sep=0pt] at (0,0) {\includegraphics[width=719.04bp]{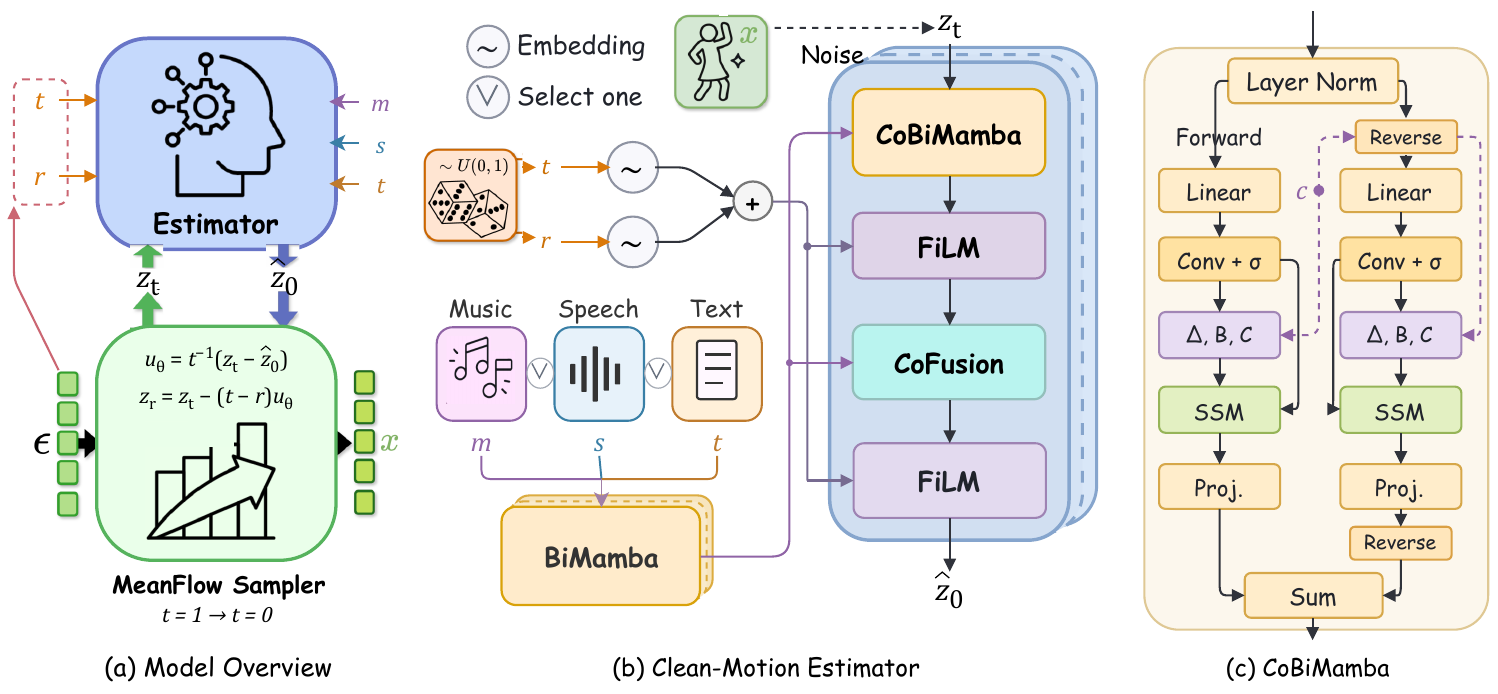}};
\fill[white] (549,309) rectangle (718,331);
\fill[fill={rgb,255:red,252;green,247;blue,237}] (622.032410,83.469963) rectangle (627.612061,98.577011);
\node[anchor=north west,inner sep=0pt] at (622.032410,79.174663) {\includegraphics[trim=622.032410bp 234.542984bp 91.427917bp 83.469963bp,clip,width=5.579651bp]{figs/FlashMotion-Overview-20260922-v5.pdf}};
\node[anchor=base,inner sep=0pt,font=\sffamily\fontsize{11.7}{13}\selectfont] at (627.53,324.8) {(c) BiMamba / CoBiMamba};
\end{tikzpicture}}
    \vspace{-8pt}
\caption{Overview of \textbf{FlashMotion}. (a) Inference flow with clean-motion prediction and MeanFlow updates. (b) Training model inputs and output: noisy motion, conditions, and timesteps are mapped to the clean-motion target. (c) Shared BiMamba/CoBiMamba block. Dashed purple paths denote optional conditioning, omitted in BiMamba and used in CoBiMamba for condition-guided state selection.}
    \label{fig:overview}
\end{figure*}

\subsection{3D Motion Generation}
3D human motion generation has made substantial progress across text-to-motion, speech-to-gesture, and music-to-dance tasks. We review existing methods from 2 complementary perspectives: \textbf{generative strategy}, which determines how motion is synthesized, and \textbf{model architecture}, which determines how temporal dependencies and conditioning signals are modeled.

\textbf{Generative Strategy.}
Representative strategies include:
(1) \textit{Direct Regression}, which directly predicts motion from conditioning signals, including Language2Pose~\cite{ahuja2019language2pose} for text-to-motion, Gesticulator~\cite{kucherenko2020gesticulator} for speech-to-gesture, and FACT~\cite{li2021ai} for music-to-dance;
(2) \textit{Adversarial Learning}, which encourages realistic motion through discriminator-based training, including the trimodal gesture model~\cite{yoon2020speech} for speech-to-gesture, and Choreography cGAN~\cite{huang2021choreography}, Dancing to Music~\cite{lee2019dancing}, and CoheDancers~\cite{yang2025cohedancers} for music-driven dance generation;
(3) \textit{Autoregressive Generation}, which synthesizes motion sequentially, including T2M-GPT~\cite{zhang2023generating} for text-to-motion, and Dance Revolution~\cite{huang2020dance}, Bailando~\cite{siyao2022bailando}, and MEGA~\cite{yang2025megadance} for music-to-dance, as well as CoDancers~\cite{yang2024codancers} for group choreography;
(4) \textit{Diffusion}, which models conditional motion distributions through iterative denoising~\cite{yang2026mace}, including MDM~\cite{tevet2023human} and MLD~\cite{chen2023executing} for text-to-motion, DiffGesture~\cite{zhu2023taming} and PersonaGesture~\cite{zhang2026personagesture} for speech-to-gesture, EDGE~\cite{tseng2023edge} and CustomDance~\cite{tang2026customdance} for dance generation and editing, and BiTDiff~\cite{jia2026bitdiff} for conducting motion; and
(5) \textit{Flow Matching}, which learns continuous transformations from noise to motion, including Motion Flow Matching~\cite{hu2023motion} and FlowMotion~\cite{canalescuba2025flowmotion} for text-driven motion, the unified speech-gesture synthesis framework~\cite{mehta2024unified}, and FlowerDance~\cite{yang2025flowerdance} for few-step music-to-dance generation.

\textbf{Model Architecture.}
Representative temporal backbones include:
(1) \textit{Recurrent Networks}, which capture temporal dependencies through sequential hidden-state updates, as exemplified by ERD~\cite{fragkiadaki2015recurrent} and the residual RNN~\cite{martinez2017human} for human motion prediction, and the recurrent decoder in Dance Revolution~\cite{huang2020dance} for music-to-dance;
(2) \textit{Transformers}, which use attention for temporal and conditional modeling, including T2M-GPT~\cite{zhang2023generating} for text-to-motion, DiffGesture~\cite{zhu2023taming} and StreamTalk~\cite{zhang2026streamtalk} for speech-to-gesture, and FACT~\cite{li2021ai} and EDGE~\cite{tseng2023edge} for music-to-dance; and
(3) \textit{Mamba-Based Models}, which exploit selective state-space modeling for efficient sequence processing, including Motion Mamba~\cite{zhang2024motionmamba} for text-to-motion, MambaTalk~\cite{xu2024mambatalk} and MambaGesture~\cite{fu2024mambagesture} for speech-to-gesture, and FlowerDance~\cite{yang2025flowerdance}, TokenDance~\cite{yang2026tokendance}, and the hybrid Mamba-Transformer model MATHDance~\cite{yang2025mathdance} for music-to-dance.

However, high-quality long-horizon generation under a strict inference budget remains challenging due to 2 coupled bottlenecks: (1) generative strategies that require many sampling steps for high-quality 3D motion generation and (2) model architectures with substantial per-step computational and memory costs for long sequences.
These costs leave limited computational resources for other production stages, such as rigging, retargeting, and rendering, potentially degrading the user experience in interactive applications.

\section{Methodology}

\subsection{Overview}
Since condition encoding, motion retargeting, and rendering also consume the end-to-end latency budget, motion generation must remain highly efficient.
For example, interactively creating and revising an extended dance or action sequence requires rapid generation of long, coherent motions so that users can promptly preview and refine the results.
Ultra-fast long-term 3D motion generation that preserves physical plausibility, conditional alignment, and motion expressiveness is therefore essential for responsive avatar animation.

Given a conditioning sequence $C=(c_i)_{i=1}^{N}$, our objective is to generate a corresponding 3D motion sequence $M=(m_i)_{i=1}^{N}$, where $c_i\in\mathbb{R}^{d_c}$ and $m_i\in\mathbb{R}^{d_m}$ denote the condition and motion features at frame $i$, respectively. We consider text, speech, and music conditioning, with a separate model trained for each task.
\textbf{(1) Text-to-Motion.}
For text description $q_k$ and its frame interval $I_k$, $c_i=E_{\mathrm{text}}(q_k)\in\mathbb{R}^{4096}$ uses UMT5-XXL~\cite{chung2023unimax}; $m_i\in\mathbb{R}^{263}$ follows HumanML3D~\cite{guo2022generating} at 20 FPS.
\textbf{(2) Speech-to-Gesture.}
The condition $c_i$ combines frame-aligned 1024-dimensional HuBERT speech features and 300-dimensional text features; $m_i\in\mathbb{R}^{333}$ comprises local 6D rotations~\cite{zhou2019continuity} for 55 SMPL-X joints and 3 root-motion components at 30 FPS.
\textbf{(3) Music-to-Dance.}
$c_i\in\mathbb{R}^{35}$ contains frame-aligned Librosa music features~\cite{mcfee2015librosa}; $m_i\in\mathbb{R}^{147}$ comprises 3D root motion (horizontal displacements and height) and 24-joint 6D rotations~\cite{zhou2019continuity} at 30 FPS.

To meet this strict inference budget in practice, we co-design \textbf{FlashMotion} across both the generative strategy and model architecture, as summarized in Fig.~\ref{fig:overview}.
(1) At the generative-strategy level, we formulate MeanFlow in 3D motion space by directly predicting clean motion and deriving the interval-averaged velocity for few-step sampling, with auxiliary kinematic and dynamic constraints to preserve motion fidelity.
(2) At the model-architecture level, we develop a lightweight non-autoregressive BiMamba backbone with condition-guided state selection and AdaIN-based condition fusion mechanism, enhancing cross-modal alignment and generalization to longer motion sequences.
(3) To support real-time human-in-the-loop motion authoring and responsive streaming applications under few-step sampling, we further introduce an Adaptive Soft masking strategy that enables coherent motion editing without additional training.

\subsection{Generative Strategy}
\noindent\textbf{MeanFlow.}
From a kinematic perspective, 3D human motion consists of continuous trajectories of root translation and joint rotations, coupled through skeletal structure and temporal dependencies~\cite{seth2018opensim,flash1985coordination}. Generating coherent motion therefore requires jointly modeling spatial coordination and temporal evolution, motivating continuous refinement of the entire sequence from unstructured noise toward a condition-consistent trajectory.
Flow Matching (FM)~\cite{lipman2022flow} learns a velocity field for an ordinary differential equation (ODE) transporting Gaussian noise $\epsilon\sim\mathcal N(0,I)$ to motion $x\sim p_{\rm data}$. Training uses the linear conditional path $z_t=\mathrm{FP}(x,t)$, with time $t\sim U(0,1)$:
\begin{equation}
\begin{aligned}
\text{FP}(x, t) = (1-t)x + t\epsilon
\label{eq: FP}
\end{aligned}
\end{equation}
The conditional path velocity is $v_{\rm cond}=\epsilon-x$. We regress $v_\theta$ against this sample-wise target; the optimal regressor estimates the marginal velocity field $v(z_t,t)=\mathbb E[v_{\rm cond}\mid z_t,t]$:
\begin{equation}
\begin{aligned}
\mathcal{L}_{\text{FM}} &= \mathbb{E}\big[\|v_\theta(z_t,t) - v_{\rm cond}\|^2\big].
\end{aligned}
\end{equation}
At inference, generation starts from an initial prior sample \(z_{1}\) and iteratively integrates the learned velocity field \(v_\theta(z_t, t)\) until reaching \(z_{0}\) in the data distribution:
\begin{equation}
\begin{aligned}
v_\theta(z_t,t) &= \frac{d z_t}{dt}, \\
z_0 &= z_1 + \int_1^0 v_\theta(z_\tau,\tau)\,d\tau.
\end{aligned}
\end{equation}

Even when conditional flows are designed to be straight, latent trajectories in the high-dimensional space are often highly curved. When using large integration intervals, Flow Matching~\cite{lipman2022flow}, which models the instantaneous velocity field, cannot accurately capture these curved trajectories, leading to degraded generation quality. Thus, \textbf{MeanFlow}~\cite{geng2025mean} replaces instantaneous velocity $v(z_t, t)$ with the interval mean velocity $u(z_t, r, t)$, leading to a training objective aligned with the actual ODE inference procedure:
\begin{equation}
u(z_t, r, t) = \frac{1}{t-r} \int_r^t v(z_\tau, \tau)\, d\tau,
\label{eq: meanflow}
\end{equation}
where $(r,t)$ are 2 timesteps randomly sampled from a uniform distribution $U(0,1)$ with the constraint $t > r$. Differentiating Eq. \ref{eq: meanflow} with respect to $t$ yields the MeanFlow identity:
\begin{equation}
u(z_t, r, t) = v(z_t,t) - (t-r)\frac{d}{dt}u(z_t, r, t).
\end{equation}
A velocity estimator $u_\theta$ is trained to satisfy the MeanFlow identity:
\begin{equation}
\label{eq:meanflow_objective}
\begin{aligned}
\mathcal{L}_{\text{MF}}
&= \mathbb{E}\big[\|u_\theta(z_t,r,t) - \text{sg}(u_{\text{tgt}})\|_2^2\big], \\
\text{where} \quad u_{\text{tgt}} &= v_{\rm cond} - (t-r)D_tu_\theta, \\
D_tu_\theta &= \partial_z u_\theta\,v_{\rm cond}+\partial_tu_\theta.
\end{aligned}
\end{equation}
where $\text{sg}(\cdot)$ denotes stop-gradient. Notably, this reduces to the FM loss when $r=t$. During inference, MeanFlow uses a constant interval-averaged velocity for each update. The general update and its equivalent integral form over the full interval are:
\begin{equation}
\begin{aligned}
z_r &= z_t - (t-r)u(z_t,r,t),
\end{aligned}
\label{eq: meanflow-sampling}
\end{equation}

\noindent\textbf{Training.}
Although MeanFlow has shown promising results in image generation~\cite{geng2025mean}, video rendering~\cite{park2026flashrender}, and audio synthesis~\cite{li2026meanaudio}, directly applying it to few-step 3D motion generation is non-trivial. 
A key reason is that these modalities often exhibit explicit local continuity in their native representations, such as spatially neighboring pixels, adjacent video frames, or temporally continuous waveforms, which provides a relatively direct structure for learning smooth generative fields. 
In contrast, SMPL-based 3D motion is represented as a sequence of pose parameters whose temporal continuity is explicit across frames, while the intra-frame dependencies among joints are only implicitly encoded by the human kinematic chain. 
Under large integration intervals, small errors in the predicted velocity field can therefore accumulate over time and disrupt these latent kinematic constraints, leading to temporal jitter, global drift, foot sliding, and implausible articulations. 
Moreover, the vanilla MeanFlow objective only constrains the averaged velocity field, without explicitly regularizing the recovered motion in kinematic or physical spaces.

To enable direct motion-level supervision, \textbf{FlashMotion} predicts the clean-motion estimate $\hat z_0$ from the current state $z_t$, frame-aligned conditioning sequence $C$, and time interval $(r,t)$.
The interval-averaged velocity is then derived as:
\begin{equation}
u_\theta(z_t,r,t;C)=\frac{z_t-\hat z_0}{t}.
\label{eq:motion_meanflow_parameterization}
\end{equation}
We optimize the induced velocity using the MeanFlow objective in Eq.~\eqref{eq:meanflow_objective}, with JVP-based differentiation, $t^2$ time weighting, and adaptive loss weighting.

Empirically, training with $\mathcal{L}_{\mathrm{MF}}$ alone still remains unstable and fails to converge in our SMPL-based experiments, as corroborated by the results in Sec.~\ref{sec:ablation_study}.
We therefore add reconstruction supervision to anchor the clean-motion estimate to ground truth:
\begin{equation}
\mathcal{L}_{\mathrm{rec}}
=\mathbb{E}\big[\|\hat z_0-z_0\|_2^2\big].
\end{equation}

To improve kinematic consistency, we match the recovered 3D joint positions to ground truth, encouraging spatially coordinated poses:
\begin{equation}
\mathcal{L}_{\mathrm{kin}}=\mathbb{E}\big[\|\hat P-P\|_2^2\big].
\end{equation}
Here, $\hat P=\mathcal{K}(\hat z_0)$ and $P=\mathcal{K}(z_0)$, where $\mathcal{K}$ recovers joint positions from 3D rotation representations.

To improve temporal dynamics, we match joint velocities and accelerations to ground truth, encouraging temporally coherent motion evolution:
\begin{equation}
\mathcal{L}_{\mathrm{dyn}}=\mathbb{E}\big[\|\Delta\hat P-\Delta P\|_2^2+\beta\|\Delta^2\hat P-\Delta^2P\|_2^2\big].
\end{equation}
Here, $\Delta$ denotes frame-wise finite differences, and $\beta$ balances velocity and acceleration supervision.

Overall, the shared training objective is:
\begin{equation}
\mathcal{L}=\lambda_{\mathrm{mf}}\mathcal{L}_{\mathrm{MF}}+\lambda_{\mathrm{rec}}\mathcal{L}_{\mathrm{rec}}+\lambda_{\mathrm{kin}}\mathcal{L}_{\mathrm{kin}}+\lambda_{\mathrm{dyn}}\mathcal{L}_{\mathrm{dyn}}.
\end{equation}

\noindent\textbf{Inference.}
Starting from Gaussian noise $\epsilon$, we progressively generate motion over uniformly spaced time intervals.
At each step, we predict the clean motion $\hat z_0$ and obtain $z_r$ by linearly interpolating between $\hat z_0$ and the current noisy state $z_t$.
This update is equivalent to Eq.~\ref{eq: meanflow-sampling}, without explicit velocity computation or additional noise injection.

\noindent\textbf{Motion Editing.}
Extending the inpainting idea in diffusion-based motion generation~\cite{liu2025gcdance,tseng2023edge}, \textbf{FlashMotion} supports training-free constraint-based editing with multiple sampling steps; 1-step editing is not supported.
This capability serves 2 complementary uses.
\textbf{(1) Human-in-the-loop motion authoring.} Users can fill temporal gaps or modify selected joint trajectories, such as interpolating between dance clips or adjusting hand paths (Fig.~\ref{fig:motionedition}).
\textbf{(2) Responsive Streaming Applications.} Temporal inpainting enables sequential window extension by constraining each new window with the previous window's tail.
This allows \textbf{FlashMotion} to adapt to incoming conditions for conversational avatars or interactive NPCs.
With few sampling steps, fixed hard masks may cause dynamic mismatches at constraint boundaries.
We therefore apply an Adaptive Soft mask, gradually relaxing the influence of known regions to improve coupling between constrained and unconstrained motions.
Given a joint-wise or temporal constraint \(x' \in \mathbb{R}^{N \times d_m}\) with positions indicated by a binary mask  \(M \in \{0,1\}^{N \times d_m}\) at time $t$, where $N$ is the sequence length and $d_m=263$, $333$, and $147$ for text-to-motion, speech-to-gesture, and music-to-dance, respectively, \textbf{FlashMotion} performs constraint-aware denoising as follows:
\begin{equation}
\begin{aligned}
z_{t} := (t \cdot M) \odot \mathrm{FP}(x', t) 
+ (1 - (t \cdot M)) \odot z_{t},
\label{eq:editing_sampling}
\end{aligned}
\end{equation}
where \(\odot\) denotes the Hadamard product, and known regions follow the constraint flow.

\begin{table*}[!t]
\centering
\begingroup
\centering
\caption{Quantitative comparison on the BABEL~\cite{punnakkal2021babel} dataset for long-sequence text-to-motion generation. Bold and underlined values denote the best and second-best generated results, excluding ground truth.}
\label{tab:babel_mixed}
\vspace{-8pt}
\setlength{\tabcolsep}{3.5pt}
\renewcommand{\arraystretch}{1.22}
\fontsize{8}{10}\selectfont
\resizebox{0.93\textwidth}{!}{\begin{tabular}{@{}ll|rrrr|rr|rr|rr@{}}
\toprule
\multicolumn{1}{@{}c}{Method} & \multicolumn{1}{c|}{Venue} & \multicolumn{1}{c}{R@1$\uparrow$} & \multicolumn{1}{c}{R@2$\uparrow$} & \multicolumn{1}{c}{R@3$\uparrow$} & \multicolumn{1}{c|}{MM$\downarrow$} & \multicolumn{1}{c}{FID$\downarrow$} & \multicolumn{1}{c|}{$\Delta$Div$\downarrow$} & \multicolumn{1}{c}{$\Delta$PJ$\downarrow$} & \multicolumn{1}{c|}{$\Delta\mathrm{AUJ}\downarrow$} & \multicolumn{1}{c}{FPS@1K$\uparrow$} & \multicolumn{1}{c@{}}{FPS@10K$\uparrow$} \\
\midrule
Ground Truth & -- & 28.504 & 43.900 & 53.161 & 4.316 & -- & 0.000 & 0.00000 & 0.00000 & -- & -- \\
MDM~\cite{tevet2023human} & ICLR'23 & 23.706 & 38.571 & 48.695 & 4.655 & 1.374 & 0.129 & 0.36381 & 1.05195 & 284 & 299 \\
ST2M~\cite{li2023st2m} & ICCV'23 & 25.793 & 39.681 & 48.900 & \underline{4.415} & 1.450 & 0.589 & 0.25851 & 1.56450 & 2\,772 & 3\,417 \\
DoubleTake~\cite{shafir2024prior} & ICLR'24 & 18.894 & 29.861 & 37.929 & 4.836 & 1.805 & 0.918 & 0.28694 & 1.00752 & 185 & 180 \\
FlowMDM~\cite{barquero2024seamless} & CVPR'24 & 19.483 & 30.783 & 39.033 & 4.847 & \underline{0.712} & 0.173 & 0.07197 & 0.27793 & 24 & 22 \\
MoMask~\cite{guo2024momask} & CVPR'24 & 25.259 & 39.390 & 49.751 & 4.546 & 1.101 & 0.271 & 0.80482 & 1.79723 & 941 & 1\,014 \\
MMM~\cite{pinyoanuntapong2024mmm} & CVPR'24 & \underline{26.936} & 39.533 & 47.709 & 4.612 & 1.823 & \underline{0.034} & 0.91381 & 1.97710 & 1\,332 & 1\,350 \\
BAMM~\cite{pinyoanuntapong2024bamm} & ECCV'24 & 26.849 & \underline{40.139} & \underline{49.855} & 4.487 & 1.270 & 0.230 & 0.77382 & 1.75389 & 144 & 152 \\
DART~\cite{zhao2025dartcontrol} & ICLR'25 & 26.105 & 39.924 & 48.485 & 4.424 & 1.241 & 0.261 & 0.05516 & 0.33090 & 75 & 75 \\
MotionStreamer~\cite{xiao2025motionstreamer} & ICCV'25 & 23.676 & 37.492 & 46.842 & 4.762 & 2.216 & \textbf{0.023} & 0.01293 & \underline{0.14735} & 10 & 10 \\
FloodDiffusion~\cite{cai2026flooddiffusion} & CVPR'26 & 25.109 & 38.924 & 48.372 & 4.491 & \textbf{0.542} & 0.199 & 0.12539 & 1.18581 & 34 & 35 \\
\midrule
\textbf{FlashMotion} (1-step) & Ours & 26.371 & 40.083 & 48.851 & 4.456 & 1.630 & 0.093 & \underline{0.00663} & 0.21474 & \textbf{29\,767} & \textbf{173\,172} \\
\textbf{FlashMotion} (4-step) & Ours & \textbf{29.134} & \textbf{42.829} & \textbf{51.683} & \textbf{4.340} & 0.907 & 0.210 & \textbf{0.00027} & \textbf{0.01295} & \underline{10\,711} & \underline{46\,303} \\
\bottomrule
\end{tabular}}
\endgroup
\par\vspace{2pt}
\begin{minipage}[t]{\dimexpr(\textwidth-\columnsep)/2\relax}
\vspace{0pt}
\begingroup
\centering
\renewcommand{\arraystretch}{1.25}
\setlength{\tabcolsep}{4pt}
\small
\caption{Quantitative comparison on FineDance~\cite{li2023finedance}.}
\label{tab:comparison_finedance_main}
\vspace{-8pt}
\resizebox{0.95\columnwidth}{!}{\begin{tabular}{ll|cc|cc|c|r}
\toprule
Method & Venue & FID$_{k}\downarrow$ & FID$_{g}\downarrow$ & DIV$_{k}\uparrow$ & DIV$_{g}\uparrow$ & BAS$\uparrow$ & FPS@10K$\uparrow$ \\
\midrule
Ground Truth & -- & -- & -- & 9.94 & 7.54 & 0.201 & -- \\
FACT~\cite{li2021ai} & ICCV'21 & 113.38 & 97.05 & 3.36 & 6.37 & 0.183 & 27 \\
MNET~\cite{kim2022brand} & CVPR'22 & 104.71 & 90.31 & 3.12 & 6.14 & 0.186 & 2\,311 \\
Bailando~\cite{siyao2022bailando} & CVPR'22 & 82.81 & 28.17 & 7.74 & 6.25 & 0.202 & 673 \\
EDGE~\cite{tseng2023edge} & CVPR'23 & 94.34 & 50.38 & 8.13 & 6.45 & 0.212 & 589 \\
FineNet~\cite{li2023finedance} & ICCV'23 & 65.15 & 23.81 & 5.84 & 5.19 & 0.219 & 1\,137 \\
Lodge~\cite{li2024lodge} & CVPR'24 & 50.00 & 35.52 & 5.67 & 4.96 & 0.226 & 884 \\
MEGA~\cite{yang2025megadance} & NeurIPS'25 & 50.00 & \textbf{13.02} & 6.23 & 6.27 & 0.226 & 340 \\
FlowerDance~\cite{yang2025flowerdance} & ECCV'26 & 29.73 & \underline{19.59} & 8.42 & \underline{7.18} & 0.232 & 4\,145 \\
\midrule
\textbf{FlashMotion} (1-step) & Ours & \underline{18.32} & 28.48 & \underline{9.48} & \textbf{7.54} & \underline{0.237} & \textbf{159\,800} \\
\textbf{FlashMotion} (4-step) & Ours & \textbf{15.15} & 21.13 & \textbf{9.55} & 6.58 & \textbf{0.240} & \underline{40\,545} \\
\bottomrule
\end{tabular}}
\vspace{-3pt}
\endgroup
\end{minipage}\hfill
\begin{minipage}[t]{\dimexpr(\textwidth-\columnsep)/2\relax}
\vspace{0pt}
\begingroup
\centering
\caption{Speech-to-gesture comparison on BEAT2~\cite{liu2024emage}.}
\label{tab:comparison_beat2_full55}
\vspace{-8pt}
\setlength{\tabcolsep}{2.5pt}
\renewcommand{\arraystretch}{1.2}
\fontsize{8}{10}\selectfont
\resizebox{0.95\columnwidth}{!}{\begin{tabular}{@{}ll|rrr|r@{}}
\toprule
Method & Venue & FGD$\downarrow$ & BC$\uparrow$ & DIV$\uparrow$ & FPS@10K$\uparrow$ \\
\midrule
Ground Truth & -- & -- & 0.4445 & 8.3714 & -- \\
CaMN~\cite{liu2022beat} & ECCV'22 & 0.5120 & 0.2000 & 5.5800 & 7\,955 \\
Audio2Photoreal~\cite{ng2024audio2photoreal} & CVPR'24 & 0.8490 & 0.3260 & 6.2400 & 115 \\
EMAGE~\cite{liu2024emage} & CVPR'24 & 0.6920 & 0.2840 & 6.0600 & 1\,839 \\
RAG-Gesture~\cite{mughal2025raggesture} & CVPR'25 & 0.4870 & \underline{0.5140} & \textbf{9.9400} & 16 \\
SemTalk~\cite{zhang2025semtalk} & ICCV'25 & \underline{0.3560} & 0.5100 & 8.4090 & 358 \\
EchoMask~\cite{zhang2025echomask} & MM'25 & 0.5660 & 0.4950 & \underline{9.2990} & 510 \\
\midrule
\textbf{FlashMotion} (1-step) & Ours & 0.4756 & 0.4801 & 6.8688 & \textbf{135\,179} \\
\textbf{FlashMotion} (4-step) & Ours & \textbf{0.3242} & \textbf{0.6006} & 9.1950 & \underline{34\,655} \\
\bottomrule
\end{tabular}}
\vspace{-3pt}
\endgroup
\end{minipage}
\end{table*}

\subsection{Model Architecture}
\noindent\textbf{Overview.}
Given the flow-path state $z_t$, conditioning sequence $C$, and interval $(r,t)$, \textbf{FlashMotion} predicts a clean-motion estimate $\hat{z}_0$, which is converted into the interval-averaged velocity for the ODE solver (Fig.~\ref{fig:overview}).
Condition features are projected into a shared hidden dimension and processed by stacked BiMamba blocks. For music and speech, genre and speaker embeddings, respectively, provide additional style guidance through gated modulation~\cite{arevalo2017gated}.
The current time and interval length $t,t-r$ are separately encoded using sinusoidal embeddings and MLPs, then combined by addition.
The motion estimator comprises stacked blocks integrating condition-guided BiMamba (CoBiMamba) for temporal modeling, FiLM~\cite{perez2018film} for time conditioning, and AdaIN-based condition fusion (CoFusion) ~\cite{huang2017arbitrary} for channel-wise condition modulation.
For long-term generation, \textbf{FlashMotion} jointly predicts all motion frames in a non-autoregressive pass at each sampling step.
These components work synergistically to enable \textbf{FlashMotion} to generate high-quality motion efficiently.

\noindent\textbf{Bidirectional Mamba.}
Our choice of Mamba is motivated by the following considerations: (1) Human motion requires strong local temporal continuity. Standard self-attention without positional information is permutation-equivariant; positional encodings provide sequence-order information~\cite{vaswani2017attention}. Owing to its inherent sequential inductive bias, Mamba~\cite{gu2023mamba} has demonstrated strong performance in modeling fine-grained local dependencies ~\cite{xu2024mambatalk,fu2024mambagesture}. (2) Mamba has only $\mathcal{O}(n)$ computational complexity compared to the $\mathcal{O}(n^2)$ complexity of attention‑based Transformers~\cite{vaswani2017attention}. (3) Mamba supports non‑autoregressive generation, which not only further improves efficiency but also mitigates the exposure‑bias problem in autoregressive methods~\cite{siyao2022bailando,yang2025megadance} and alleviates transition artifacts at segment boundaries in inpainting‑based approaches~\cite{tseng2023edge,liu2025gcdance}. 

Furthermore, condition--motion dependencies benefit from bidirectional temporal context, but standard Mamba models sequences only in a single direction. To address this limitation, we introduce Bidirectional Mamba (BiMamba) and condition-guided BiMamba (CoBiMamba), which capture information from both temporal directions. As shown in Fig.~\ref{fig:overview}, the sequence is processed by forward and backward Mamba branches; their outputs are fused by addition and refined through a multiplicative skip connection, enhancing gradient flow and feature preservation. This bidirectional design improves motion coherence and strengthens condition--motion alignment.

\noindent\textbf{Condition-Guided State Selection.}
Mamba is built upon state space models (SSMs), which capture temporal dependencies through recurrent hidden-state updates. Specifically, its selective SSM (S6) uses input-dependent parameters to regulate information retention and propagation, while a hardware-aware parallel scan enables efficient computation~\cite{gu2023mamba}.
Unlike S4~\cite{gu2021efficiently}, whose learned SSM coefficients are invariant across sequence positions, Mamba makes $B_i$, $C_i$, and $\Delta_i$ input-dependent through shared learned projections, while retaining a shared, input-independent learnable matrix $A$.
The core SSM recurrence at frame $i$ is:
\begin{equation}
\begin{aligned}
h_i &= \bar{A}_i h_{i-1} + \bar{B}_i x_i, \\
y_i &= C_i h_i.
\end{aligned}
\end{equation}
Here, $i$ indexes motion frames, distinct from the generative time $t$; the block's skip connection, gating, and output projection are omitted from this core formulation.
Following the Mamba implementation, we use a diagonal $A$, an input-dependent discretization step $\Delta_i$, and simplified input discretization:
\begin{equation}
\begin{aligned}
\bar{A}_i &= \exp(\Delta_i A), \\
\bar{B}_i &= \Delta_i B_i.
\end{aligned}
\end{equation}
These operations apply independently per channel, with $B_i$ and $C_i$ shared across channels.
A hardware-aware parallel scan efficiently evaluates the recurrence, using projection weights and $A$ shared across frames.

For condition encoding, BiMamba derives its selection coefficients from its own condition features.
In the motion backbone, CoBiMamba jointly uses locally processed motion features $\tilde X_i$ and aligned, time-modulated condition features $G_i$:
\begin{equation}
\begin{aligned}
B_i &= f_B(\tilde X_i)+g_B(G_i), \quad C_i=f_C(\tilde X_i)+g_C(G_i),\\
\Delta_i &= \operatorname{softplus}\!\big(W_\Delta[f_\Delta(\tilde X_i)+g_\Delta(G_i)]+b_\Delta\big).
\end{aligned}
\end{equation}
Here, $\tilde X_i$ denotes motion features after input projection, local convolution, and activation.
The learned projections $f$ and $g$ are shared across frames; $f_\Delta$ and $g_\Delta$ produce low-dimensional intermediate features expanded by the same $W_\Delta$, with a learned bias $b_\Delta$.
$A$ remains condition-independent.
Conditioning thus regulates memory retention through $\Delta_i$, information injection through $B_i$, and state readout through $C_i$, enabling motion-state updates to follow changing action instructions or rhythmic cues.

\noindent\textbf{Channel-Level Condition Fusion.}
To integrate text, speech, or music conditions efficiently, we use AdaIN-based residual modulation~\cite{huang2017arbitrary}. Frame-wise condition embeddings predict bounded scale and shift residuals for temporally normalized motion features, which are rescaled with their original statistics. Zero-initialized modulation preserves the identity mapping at initialization.
This design offers 3 advantages:
(1) \emph{Length-agnostic generation}: shared frame-wise projections introduce no length-specific parameters. Combined with our non-autoregressive BiMamba backbone, this allows \textbf{FlashMotion} to directly generate full long sequences even when trained on short motion windows;
(2) \emph{Efficient fusion}: for fixed feature widths, frame-wise projections and modulation have $O(N)$ computational complexity, avoiding the $O(N^2)$ pairwise interactions of cross-attention in our frame-aligned setting; and
(3) \emph{Expressive conditioning}: compared with simple additive fusion, condition-dependent scaling and shifting provide both multiplicative and additive control, enabling amplification, suppression, and offset of motion-feature channels.

\section{Experiment}

\begin{figure*}[!t]
    \centering
    \includegraphics[width=0.8\textwidth]{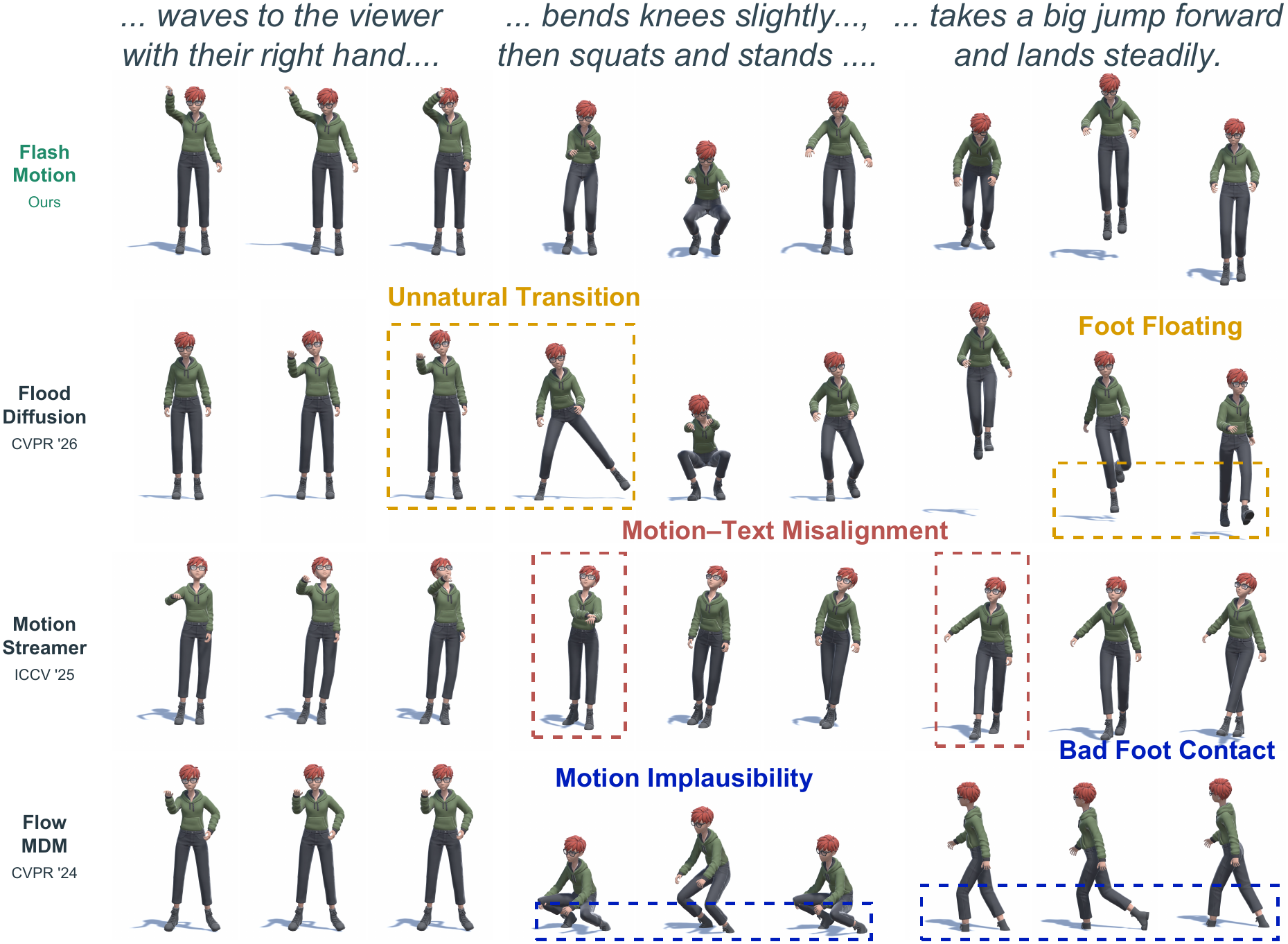}
    \vspace{-10pt}
    \caption{Text-to-motion comparison under a shared sequence of 3 prompts.}
    \label{fig:t2m_comparison}
\end{figure*}

\subsection{Experimental Setup}
\noindent\textbf{Text-to-Motion.}
\textit{(1) Dataset.}
BABEL~\cite{punnakkal2021babel} provides approximately 43 hours of motion covering over 250 action categories with temporally localized annotations; we resample motions to 20 FPS.
Following prior work~\cite{athanasiou2022teach,barquero2024seamless,zhao2025dartcontrol}, we evaluate a Mixed set of 311 validation sequences (201-1030 frames) containing approximately 300 transitions between text prompts.
\textit{(2) Evaluation.}
Following~\cite{zhao2025dartcontrol}, we assess semantic alignment using R@1/2/3 and MM, distributional quality using FID, diversity using $\Delta\mathrm{Div}$, and transition continuity using $\Delta\mathrm{PJ}$ and $\Delta\mathrm{AUJ}$.

\noindent\textbf{Music-to-Dance.}
\textit{(1) Dataset.}
We use the publicly released 7.7-hour FineDance subset~\cite{li2023finedance}, covering 16 dance genres at 30 FPS.
Following prior FineDance-based evaluations~\cite{yang2025megadance,yang2025flowerdance,yang2026tokendance}, we extract 84 overlapping clips from 19 test recordings.
\textit{(2) Evaluation.}
Following prior work~\cite{yang2025megadance,yang2025flowerdance}, we assess distributional quality using $\mathrm{FID}_{k}$ and $\mathrm{FID}_{g}$, diversity using $\mathrm{DIV}_{k}$ and $\mathrm{DIV}_{g}$, rhythmic alignment using BAS.

\noindent\textbf{Speech-to-Gesture.}
\textit{(1) Dataset.}
We use BEAT2~\cite{liu2024emage} at 30 FPS, which contains 60 hours of synchronized speech and holistic motion from 25 speakers.
Following~\cite{zhang2026streamtalk,zhang2025semtalk}, we evaluate on the 265 test recordings in the official BEAT2 English split, using each recording's full aligned speech--motion span.
\textit{(2) Evaluation.}
Following ~\cite{zhang2026streamtalk}, we assess distributional quality using FGD, speech-motion rhythmic alignment using BC, and motion diversity using DIV.

\noindent\textbf{Efficiency Evaluation.}
For our measurements, we compute generation FPS at batch size 1 across 1k or 10k output sequence lengths from the mean runtime of 5 GPU-synchronized runs after 1 full warm-up run.
Timing covers only generation between predefined input and output representations, excluding external modality encoding, retargeting, and IK.

\subsection{Quantitative Evaluation}
\noindent\textbf{Generation Quality.}
Tables~\ref{tab:babel_mixed}, \ref{tab:comparison_finedance_main}, and \ref{tab:comparison_beat2_full55} compare generation quality for text-to-motion, music-to-dance, and speech-to-gesture, respectively.
For text-to-motion, \textbf{FlashMotion} leads R@1/2/3 and MM, while reducing $\Delta\mathrm{PJ}$ and $\Delta\mathrm{AUJ}$ by 97.9\% and 91.2\% relative to the strongest external baselines.
These closer-to-ground-truth transition dynamics, improved semantic alignment, and competitive FID establish overall state-of-the-art performance in long-horizon action composition.
For music-to-dance, \textbf{FlashMotion} reduces $\mathrm{FID}_{k}$ by 49.0\% and improves $\mathrm{DIV}_{k}$ by 13.4\% over FlowerDance~\cite{yang2025flowerdance}, with a more modest 3.4\% gain in BAS.
Its 1-step variant also matches ground-truth $\mathrm{DIV}_{g}$, supporting overall state-of-the-art dance quality across motion fidelity, diversity, and rhythmic alignment.
For speech-to-gesture, \textbf{FlashMotion} achieves the lowest FGD and highest BC, improving over SemTalk~\cite{zhang2025semtalk} and RAG-Gesture~\cite{mughal2025raggesture} by 8.9\% and 16.8\%, respectively.
Combined with diversity close to ground truth, these results yield an overall state-of-the-art balance of motion fidelity, synchronization, and diversity.
Overall, \textbf{FlashMotion} combines 3D motion-oriented MeanFlow with kinematic and dynamic regularization and a condition-guided BiMamba architecture with AdaIN fusion to achieve superior 3D motion generation quality.
These advantages are strongest with 4-step inference, while even 1-step generation remains competitive across complex dance, expressive speech-driven gestures, and compositional text-driven actions, demonstrating the applicability of this design across conditioning modalities.

\noindent\textbf{Generation Efficiency.}
The FPS columns in Tables~\ref{tab:babel_mixed}, \ref{tab:comparison_finedance_main}, and \ref{tab:comparison_beat2_full55} report generation throughput.
At $10^4$ frames, 4-step \textbf{FlashMotion} achieves $13.6\times$, $9.8\times$, and $4.4\times$ the throughput of the fastest external T2M, M2D, and S2G baselines (ST2M, FlowerDance, and CaMN), respectively, with even higher throughput in 1 step.
Figure~\ref{fig: teaser} visualizes the quality--throughput trade-off across all 3 tasks.
On BABEL, 4-step throughput rises from 10,711 to 46,303 FPS as sequence length increases from $10^3$ to $10^4$ frames, consistent with efficient long-sequence processing by our non-autoregressive BiMamba backbone.
This substantial efficiency lead benefits from 2 complementary designs: \textbf{(1) 3D motion-oriented MeanFlow} enables high-quality generation in a few  steps; \textbf{(2) the Mamba-based architecture} naturally supports non-autoregressive long-sequence generation, while condition-guided state selection and AdaIN fusion further preserve motion quality, conditional alignment, and temporal continuity.

\begin{figure*}[!t]
    \centering
\vspace{0pt}
    \begin{minipage}[t]{0.48\textwidth}
    \vspace{0pt}
    \centering
    \includegraphics[width=\linewidth]{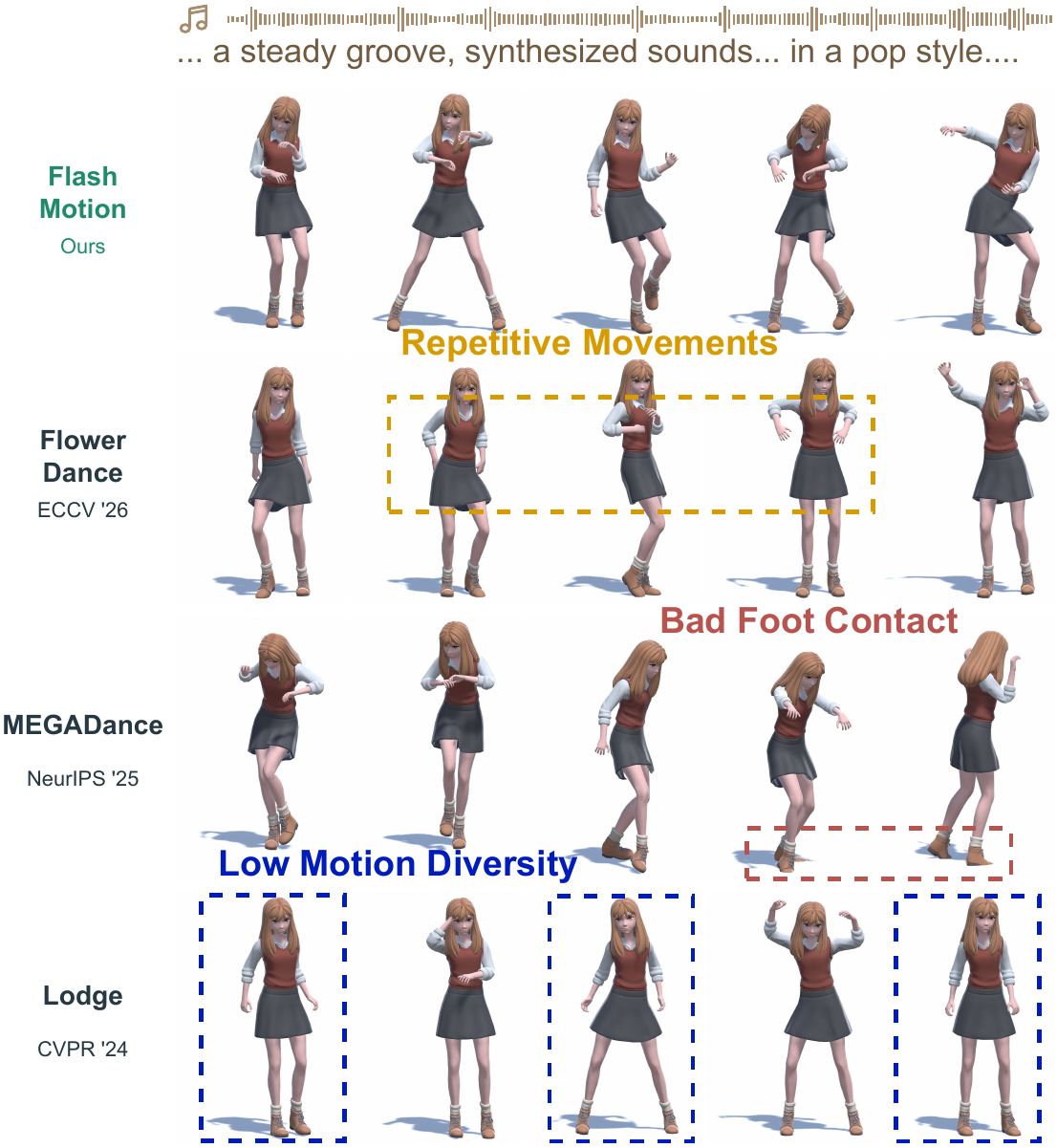}
    \vspace{-12pt}
    \caption{Music-to-dance comparison on the same music excerpt.}
    \label{fig:m2d_comparison}
    \end{minipage}\hfill
    \begin{minipage}[t]{0.48\textwidth}
    \vspace{0pt}
    \centering
    \includegraphics[width=\linewidth]{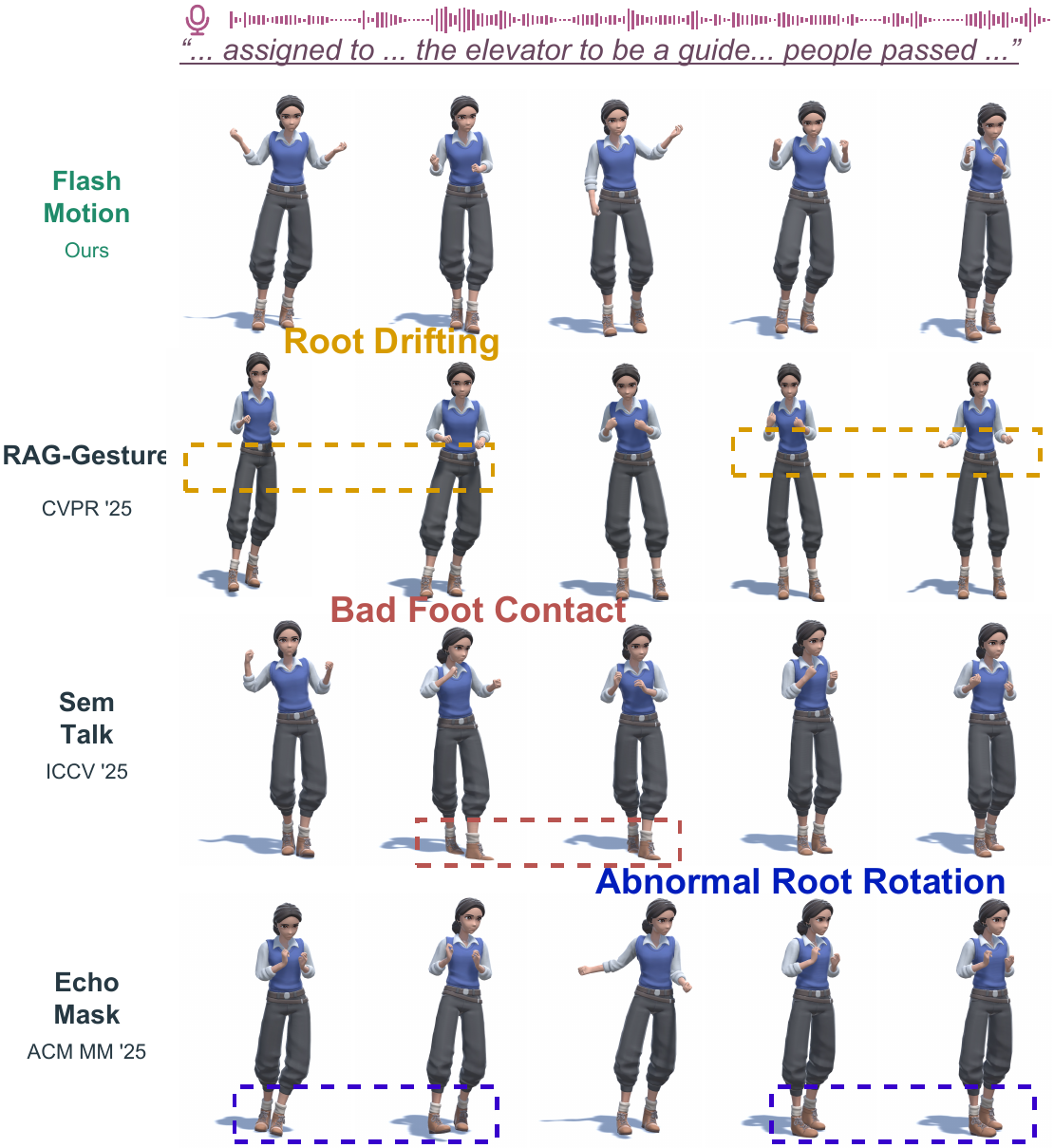}
    \vspace{-12pt}
    \caption{Speech-to-gesture comparison on the same speech excerpt.}
    \label{fig:s2g_comparison}
    \end{minipage}
\end{figure*}

\subsection{Qualitative Evaluation}
We qualitatively compare \textbf{FlashMotion} with representative baselines across 3 motion generation tasks. 
\textbf{(1) Text-to-Motion.} Each method receives 3 consecutive text prompts, each specifying a 5-second action. As shown in Fig.~\ref{fig:t2m_comparison}, \textbf{FlashMotion} follows the changing instructions while preserving articulated movements and plausible poses. In contrast, FlowMDM~\cite{barquero2024seamless} exhibits limited motion variation and unnatural joint configurations, MotionStreamer~\cite{xiao2025motionstreamer} continues similar gestures despite prompt changes, and FloodDiffusion~\cite{cai2026flooddiffusion} produces unnatural movements and abrupt transitions. 
\textbf{(2) Speech-to-Gesture.} Given the same 15-second speech clip, \textbf{FlashMotion} generates expressive gestures that follow speech rhythm while maintaining coordinated, physically plausible body movements (Fig.~\ref{fig:s2g_comparison}). RAG-Gesture~\cite{mughal2025raggesture} exhibits root drift, SemTalk~\cite{zhang2025semtalk} shows poor foot-ground contact, and EchoMask~\cite{zhang2025echomask} produces unnatural foot motion and implausible root rotations.
\textbf{(3) Music-to-Dance.} Given the same 15-second music clip, \textbf{FlashMotion} produces varied, rhythmically aligned choreography with coordinated whole-body articulation and plausible foot contact (Fig.~\ref{fig:m2d_comparison}). Lodge~\cite{li2024lodge} repeats similar movement patterns, MEGA~\cite{yang2025megadance} exhibits poor foot--floor contact, and FlowerDance~\cite{yang2025flowerdance} occasionally produces physically implausible poses. Overall, these visual comparisons demonstrate \textbf{FlashMotion}'s consistent advantages in conditional alignment, motion expressiveness, and physical plausibility across all 3 tasks. These results benefit from our complementary designs: kinematic and dynamic supervision promote plausible and smooth movements, while condition-guided state selection and AdaIN fusion align motion with diverse input conditions.\looseness=-1

\subsection{User Study}
Motion's inherent subjectivity makes user feedback essential for evaluating generated movements.
To complement objective metrics, we conduct a user study across T2M, S2G, and M2D, with 40 conditions per task and 30 human reviewers.
All reviewers provided informed consent before participating.
Ratings use a 5-point scale (1: Terrible; 2: Poor; 3: Fair; 4: Good; 5: Great) and assess \emph{Motion Quality} (MQ, plausibility and diversity), \emph{Motion Synchronization} (MS, alignment with text, speech, or music), and \emph{Motion Continuity} (MC, smooth inter-segment transitions and jitter-free motion within segments).
Table~\ref{tab:user_study} reports mean ratings, with \textbf{FlashMotion} scoring highest on every criterion across all 3 tasks.
For T2M, it exceeds the strongest baseline for each criterion by 0.16, 0.24, and 0.01 points in MQ, MS, and MC, respectively, indicating stronger perceived motion quality and text alignment with comparable continuity.
For S2G, the corresponding gains are 0.35, 0.21, and 0.14 points, showing the largest MQ improvement across tasks alongside better speech alignment and smoother gestures.
For M2D, MQ, MS, and MC reach 3.98, 3.31, and 3.99, exceeding the strongest baselines by 0.01, 0.03, and 0.04 points, respectively; these smaller margins indicate a modest but consistent advantage across the criteria.
These perceptual gains are consistent with our design: kinematic and dynamic supervision encourage plausible poses and smooth motion, while condition-guided state selection and AdaIN fusion promote cross-modal alignment, jointly improving motion quality, synchronization, and continuity.
These results support \textbf{FlashMotion}'s advantage in human preferences for motion quality, conditional alignment, and temporal continuity.
See the appendix for study details.\looseness=-1

\subsection{Ablation Study}
\label{sec:ablation_study}
Additional qualitative results appear in the appendix and demo video. We conduct ablations on text-to-motion, whose broad motion coverage and established metrics enable comprehensive evaluation of components shared across tasks.

\noindent\textbf{Generative Strategy.}
Table~\ref{tab:ablation_steps} compares MeanFlow with Flow Matching~\cite{liu2022flow} and Diffusion~\cite{tevet2023human} using 1, 2, and 4 sampling steps.
All variants use the same clean 3D motion prediction target and model architecture.
Increasing MeanFlow from 1 to 4 steps improves retrieval and FID, with closer-to-ground-truth transition dynamics.
At 4 steps, MeanFlow achieves the highest retrieval scores and lowest FID, reducing FID by 15.5\% and 18.1\% relative to Flow Matching and diffusion, respectively.
Its jerk-curve discrepancy is also 91.9\% lower than that of Flow Matching.
These results demonstrate the effectiveness of our 3D motion-oriented MeanFlow strategy for high-quality few-step generation.

\begin{table}[!t]
\centering
\vspace{0pt}
\begingroup
\centering\footnotesize
\setlength{\tabcolsep}{3pt}
\caption{User study ratings (mean $\pm$ SD; 1--5, higher is better).}
\label{tab:user_study}
\vspace{-4pt}
\begin{tabular*}{\linewidth}{@{\extracolsep{\fill}}clccc@{}}
\toprule
& Method & MQ$\uparrow$ & MS$\uparrow$ & MC$\uparrow$ \\
\midrule
\multirow{4}{*}{\rotatebox{-90}{T2M}} & DART & $3.45 \pm 0.90$ & $3.12 \pm 0.98$ & $3.65 \pm 0.82$ \\
 & FloodDiffusion & $3.62 \pm 0.89$ & $3.35 \pm 0.99$ & $3.85 \pm 0.86$ \\
 & FlowMDM & $3.62 \pm 0.86$ & $3.36 \pm 0.95$ & $3.50 \pm 0.87$ \\
 & \textbf{FlashMotion} & $\mathbf{3.78} \pm 0.86$ & $\mathbf{3.60} \pm 0.93$ & $\mathbf{3.86} \pm 0.80$ \\
\midrule
\multirow{4}{*}{\rotatebox{-90}{S2G}} & CaMN & $3.33 \pm 0.89$ & $3.02 \pm 1.00$ & $3.77 \pm 0.81$ \\
 & EMAGE & $3.42 \pm 0.92$ & $2.87 \pm 0.98$ & $3.98 \pm 0.84$ \\
 & RAG-Gesture & $3.55 \pm 0.87$ & $3.05 \pm 0.90$ & $3.88 \pm 0.81$ \\
 & \textbf{FlashMotion} & $\mathbf{3.90} \pm 0.83$ & $\mathbf{3.26} \pm 0.99$ & $\mathbf{4.12} \pm 0.78$ \\
\midrule
\multirow{4}{*}{\rotatebox{-90}{M2D}} & FlowerDance & $3.92 \pm 0.88$ & $3.27 \pm 0.98$ & $3.95 \pm 0.83$ \\
 & MEGA & $3.93 \pm 0.85$ & $3.22 \pm 0.96$ & $3.93 \pm 0.80$ \\
 & FineNet & $3.97 \pm 0.88$ & $3.28 \pm 0.97$ & $3.91 \pm 0.83$ \\
 & \textbf{FlashMotion} & $\mathbf{3.98} \pm 0.86$ & $\mathbf{3.31} \pm 0.98$ & $\mathbf{3.99} \pm 0.80$ \\
\bottomrule
\end{tabular*}
\vspace{-3pt}
\endgroup
\par\vspace{2pt}
\begingroup
\centering
\renewcommand{\arraystretch}{1.1}
\setlength{\tabcolsep}{2pt}
\captionof{table}{Generative strategy ablation on BABEL. Bold denotes the best result at each sampling budget.}
\label{tab:ablation_steps}
\vspace{-8pt}
\resizebox{\linewidth}{!}{\scriptsize
\begin{tabular}{@{}l|c|rrrrrrrr@{}}
\toprule
& S & R@1$\uparrow$ & R@2$\uparrow$ & R@3$\uparrow$ & MM$\downarrow$ & FID$\downarrow$ & $\Delta$Div$\downarrow$ & $\Delta$PJ$\downarrow$ & $\Delta\mathrm{AUJ}\downarrow$ \\
\midrule
\multirow{3}{*}{Diffusion}
& 1 & 24.622 & 36.510 & 44.539 & 4.482 & 2.166 & 0.646 & 0.00095 & 0.09766 \\
& 2 & 25.021 & 37.398 & 45.858 & 4.475 & 1.682 & 0.527 & 0.01483 & 0.24482 \\
& 4 & 25.954 & 38.570 & 46.985 & 4.433 & 1.107 & 0.471 & 0.01270 & 0.20795 \\
\midrule
\multirow{3}{*}{Flow Matching}
& 1 & 25.628 & 38.228 & 46.632 & \textbf{4.388} & 2.009 & 0.463 & \textbf{0.00024} & \textbf{0.09359} \\
& 2 & 26.394 & 39.476 & 47.931 & \textbf{4.421} & 1.302 & 0.302 & 0.01168 & 0.18384 \\
& 4 & 27.460 & 40.683 & 49.149 & 4.379 & 1.074 & 0.248 & 0.00995 & 0.16019 \\
\midrule
\multirow{3}{*}{MeanFlow}
& 1 & \textbf{26.371} & \textbf{40.083} & \textbf{48.851} & 4.456 & \textbf{1.630} & \textbf{0.093} & 0.00663 & 0.21474 \\
& 2 & \textbf{28.388} & \textbf{41.606} & \textbf{50.527} & 4.438 & \textbf{1.192} & \textbf{0.004} & \textbf{0.00539} & \textbf{0.09123} \\
& 4 & \textbf{29.134} & \textbf{42.829} & \textbf{51.683} & \textbf{4.340} & \textbf{0.907} & \textbf{0.210} & \textbf{0.00027} & \textbf{0.01295} \\
\bottomrule
\end{tabular}}
\endgroup
\par\vspace{0pt}
\centering
\captionof{table}{Motion backbone ablation on BABEL (4-step).}
\label{tab:backbone}
\vspace{-8pt}
\begingroup
\renewcommand{\arraystretch}{1.15}
\setlength{\tabcolsep}{1.4pt}
\fontsize{8}{10}\selectfont
\resizebox{\linewidth}{!}{\begin{tabular}{@{}llrrrrrrrr@{}}
\toprule
Backbone & Inference & R@1$\uparrow$ & R@2$\uparrow$ & R@3$\uparrow$ & MM$\downarrow$ & FID$\downarrow$ & $\Delta$Div$\downarrow$ & $\Delta$PJ$\downarrow$ & $\Delta\mathrm{AUJ}\downarrow$ \\
\midrule
\multirow[c]{2}{*}{Transformer}
& Non-AR & 24.805 & 37.093 & 45.400 & 4.427 & 1.937 & 0.609 & 1.01509 & 10.76431 \\
& Streamline & 25.992 & 38.953 & 47.588 & 4.341 & 1.666 & 0.625 & 1.03962 & 11.80452 \\
Mamba & Non-AR & 24.881 & 38.000 & 46.583 & 4.538 & 0.917 & 0.232 & 0.03950 & 0.63527 \\
\textbf{BiMamba} & Non-AR & \textbf{29.134} & \textbf{42.829} & \textbf{51.683} & \textbf{4.340} & \textbf{0.907} & \textbf{0.210} & \textbf{0.00027} & \textbf{0.01295} \\
\bottomrule
\end{tabular}}
\endgroup
\end{table}

\begin{table}[!t]
\centering
\vspace{0pt}
\centering
\captionof{table}{Training objective ablation on BABEL (4-step).}
\label{tab:ablation}
\vspace{-8pt}
\begingroup
\renewcommand{\arraystretch}{1.4}
\setlength{\tabcolsep}{2.8pt}
\fontsize{8}{10}\selectfont
\resizebox{\linewidth}{!}{\begin{tabular}{@{}lrrrrrrrr@{}}
\toprule
Loss setting & R@1$\uparrow$ & R@2$\uparrow$ & R@3$\uparrow$ & MM$\downarrow$ & FID$\downarrow$ & $\Delta$Div$\downarrow$ & $\Delta$PJ$\downarrow$ & $\Delta\mathrm{AUJ}\downarrow$ \\
\midrule
w/o Dynamic & 27.393 & 41.426 & 50.905 & \textbf{4.324} & 0.928 & 0.360 & 0.00097 & \textbf{0.01241} \\
w/o Kinematic & 28.613 & 42.530 & \textbf{51.813} & 4.386 & 0.983 & \textbf{0.090} & 0.00201 & 0.06503 \\
MeanFlow only & NaN & NaN & NaN & NaN & NaN & NaN & NaN & NaN \\
w/o MeanFlow & 27.460 & 40.683 & 49.149 & 4.379 & 1.074 & 0.248 & 0.00995 & 0.16019 \\
w/ Uncond & 25.456 & 38.893 & 47.860 & 4.516 & 2.232 & 0.311 & 0.00241 & 0.06472 \\
\textbf{Full Model} & \textbf{29.134} & \textbf{42.829} & 51.683 & 4.340 & \textbf{0.907} & 0.210 & \textbf{0.00027} & 0.01295 \\
\bottomrule
\end{tabular}}
\endgroup

\par\vspace{0pt}
\centering
\captionof{table}{Condition fusion and state selection ablation on BABEL (4-step).}
\label{tab:components}
\vspace{-8pt}
\begingroup
\renewcommand{\arraystretch}{1.15}
\setlength{\tabcolsep}{2.2pt}
\fontsize{8}{10}\selectfont
\resizebox{\linewidth}{!}{\begin{tabular}{@{}lrrrrrrrr@{}}
\toprule
Component & R@1$\uparrow$ & R@2$\uparrow$ & R@3$\uparrow$ & MM$\downarrow$ & FID$\downarrow$ & $\Delta$Div$\downarrow$ & $\Delta$PJ$\downarrow$ & $\Delta\mathrm{AUJ}\downarrow$ \\
\midrule
Add only & 27.208 & 40.909 & 50.120 & \textbf{4.307} & 1.073 & 0.262 & 0.00105 & 0.06135 \\
AdaIN only & 28.220 & 42.575 & 51.215 & 4.342 & 0.973 & 0.462 & 0.00101 & 0.01925 \\
Scan only & 27.849 & 42.472 & 51.349 & 4.444 & 1.616 & \textbf{0.027} & 0.00680 & 0.10705 \\
Add + Scan & 26.090 & 40.318 & 49.643 & 4.380 & 1.283 & 0.483 & 0.00152 & 0.06991 \\
\textbf{AdaIN + Scan} & \textbf{29.134} & \textbf{42.829} & \textbf{51.683} & 4.340 & \textbf{0.907} & 0.210 & \textbf{0.00027} & \textbf{0.01295} \\
\bottomrule
\end{tabular}}
\endgroup

\par\vspace{0pt}
\centering
\captionof{table}{Timestep sampling strategy ablation on BABEL (4-step).}
\label{tab:time_sampling}
\vspace{-8pt}
\begingroup
\renewcommand{\arraystretch}{1.15}
\setlength{\tabcolsep}{2pt}
\fontsize{8}{10}\selectfont
\resizebox{\linewidth}{!}{\begin{tabular}{@{}lrrrrrrrr@{}}
\toprule
Distribution & R@1$\uparrow$ & R@2$\uparrow$ & R@3$\uparrow$ & MM$\downarrow$ & FID$\downarrow$ & $\Delta$Div$\downarrow$ & $\Delta$PJ$\downarrow$ & $\Delta\mathrm{AUJ}\downarrow$ \\
\midrule
$\mathrm{Beta}(0.5,0.5)$ & 27.908 & 41.656 & 51.041 & 4.350 & \textbf{0.731} & \textbf{0.173} & 0.00216 & 0.03406 \\
$\mathrm{Beta}(2.0,2.0)$ & 24.421 & 36.698 & 45.243 & 4.535 & 2.359 & 0.363 & 0.00844 & 0.13800 \\
$\mathrm{Beta}(1.0,2.0)$ & 11.656 & 20.542 & 28.004 & 5.300 & 26.028 & 4.388 & 0.00413 & 0.08032 \\
$\mathrm{Beta}(2.0,1.0)$ & 24.221 & 36.962 & 46.106 & 4.415 & 2.685 & 0.846 & 0.01100 & 0.18491 \\
\textbf{Uniform} & \textbf{29.134} & \textbf{42.829} & \textbf{51.683} & \textbf{4.340} & 0.907 & 0.210 & \textbf{0.00027} & \textbf{0.01295} \\
\bottomrule
\end{tabular}}
\endgroup
\end{table}

\begingroup
\tolerance=800

\noindent\textbf{Training Objectives.}
Table~\ref{tab:ablation} evaluates MeanFlow-only training, removal of kinematic or dynamic regularization, and the \textit{w/o.\ MeanFlow Loss} variant under 4-step sampling.
Following MeanFlow~\cite{geng2025mean}, the \textit{w/ Uncond} variant drops conditioning with a probability of 10\% during training to jointly learn conditional and unconditional generation.
MeanFlow-only training diverges, highlighting the need for direct motion-level supervision. Training with 10\% condition dropout (w/ Uncond) increases FID from 0.907 to 2.232 and lowers R@1 from 29.134 to 25.456, indicating reduced motion fidelity and conditional alignment.
All variants use the same model architecture.
Removing kinematic supervision increases FID from 0.907 to 0.983 and $\Delta\mathrm{AUJ}$ from 0.01295 to 0.06503.
Removing dynamic supervision lowers R@1 from 29.134 to 27.393 and increases $\Delta$PJ from 0.00027 to 0.00097.
The \textit{w/o.\ MeanFlow Loss} variant also weakens motion fidelity and continuity.
The full objective achieves the best R@1/2, FID, and $\Delta$PJ, supporting the complementary benefits of generative learning, motion reconstruction, and kinematic and dynamic regularization.\looseness=-2\par
\endgroup

\begingroup
\tolerance=800

\noindent\textbf{Model Backbone.}
Table~\ref{tab:backbone} compares BiMamba, Mamba~\cite{gu2023mamba}, and Transformer~\cite{vaswani2017attention}, including its Non-AR and Streamline modes.
Non-AR directly generates long sequences after training on short windows, whereas Streamline extends motion through prefix-constrained windowed inpainting~\cite{tseng2023edge}. The Transformer Streamline variant uses positional encodings within each window, whereas the Transformer Non-AR variant omits them for direct length extrapolation. Mamba and BiMamba also omit positional embeddings. All variants use the same MeanFlow-based strategy with 4-step sampling.
BiMamba achieves the best retrieval and smallest boundary gaps, improving R@1 over Mamba by 4.25 percentage points and reducing $\Delta\mathrm{AUJ}$ by 98.0\%.
BiMamba also achieves lower FID and diversity gap than Mamba.
Both Transformer variants have much larger boundary gaps than BiMamba.
These results demonstrate BiMamba's advantage in combining strong semantic alignment with closer-to-ground-truth transition dynamics, supporting its use as the backbone of \textbf{FlashMotion}.\looseness=-1\par
\endgroup

\begin{figure*}[!t]
\centering

\begin{minipage}[t]{0.44\textwidth}
\hrule height 0pt
  \centering
  \includegraphics[width=\linewidth]{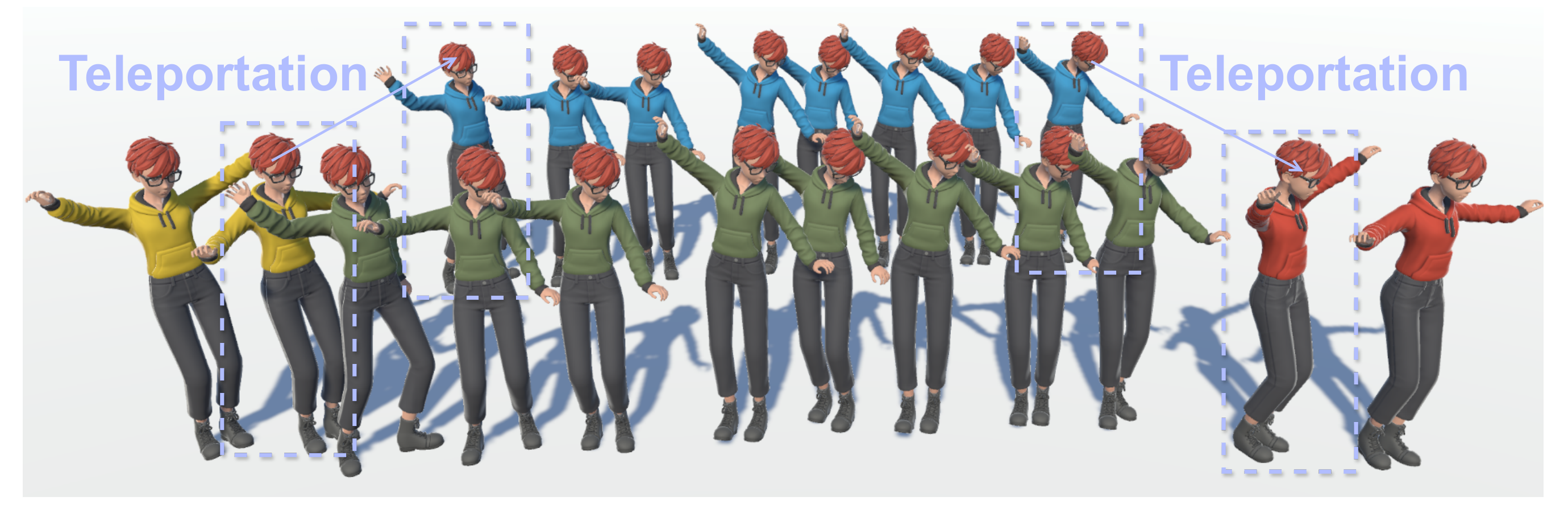}
  \par\vspace{-9pt}
  \caption{
  Given start (yellow) and end (red) motions, \textbf{FlashMotion} generates smooth in-between motions with the Adaptive Soft mask (green), whereas a hard mask (blue) causes teleportation artifacts.  For clarity, we showcase music-to-dance as a representative task.
  }
  \label{fig:motionedition}
\end{minipage}
\hfill
\begin{minipage}[t]{0.52\textwidth}
\hrule height 0pt
\centering
\vspace{-0.15in}
\captionof{table}{Quantitative evaluation of inpainting transition quality.}
\vspace{-0.18in}
\label{tab:editing_masks}
\par\vspace{4pt}
\renewcommand{\arraystretch}{1.6}
\setlength{\tabcolsep}{2.8pt}
\resizebox{\linewidth}{!}{\scriptsize
\begin{tabular}{@{}lccc@{}}
\toprule
Mask & Seam$\downarrow$ (m/s) & Jerk$\downarrow$ (m/s$^3$) & Activity ($\rightarrow1$) \\
\midrule
GT & $0.222\pm0.154$ & $155.04\pm109.66$ & $1.000\pm0.000$ \\
Hard~\cite{tevet2023human,tseng2023edge} & $2.066\pm0.834$ & $254.70\pm54.97$ & $1.337\pm0.619$ \\
Fixed Soft~\cite{shafir2024prior} & $1.657\pm0.862$ & $251.29\pm80.94$ & $1.329\pm0.619$ \\
\textbf{Adaptive Soft} & $\mathbf{0.312\pm0.271}$ & $\mathbf{243.88\pm95.38}$ & $1.327\pm0.619$ \\
\bottomrule
\end{tabular}}
\end{minipage}
\par\vspace{-2pt}
\end{figure*}

\begin{figure*}[t]
\centering
    
    \includegraphics[width=\textwidth]{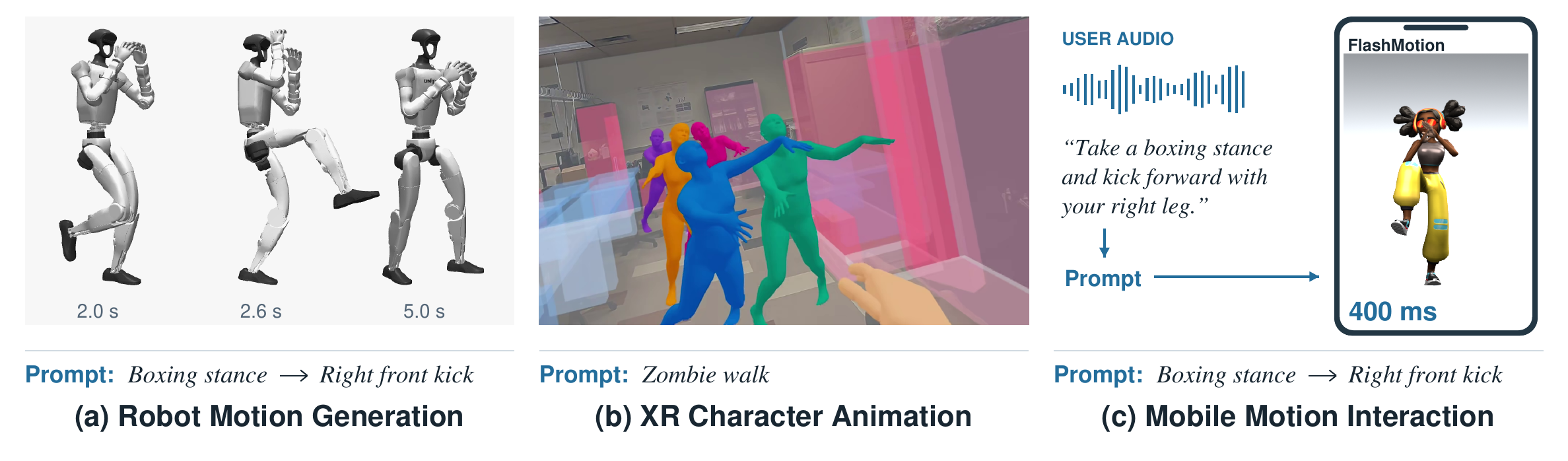}
    \vspace{-24pt}
\caption{With lightweight post-processing, \textbf{FlashMotion} supports various downstream applications, including (a) robot motion generation, (b) XR character animation, and (c) mobile motion interaction. For clarity, we showcase text-to-motion as a representative task.}
    \label{fig:applications}
\end{figure*}

\begingroup
\tolerance=800
\noindent\textbf{Model Components.}
Table~\ref{tab:components} compares additive fusion, AdaIN~\cite{huang2017arbitrary}, condition-guided state selection (Scan), and their combinations under 4-step sampling.
Replacing addition with AdaIN improves retrieval and FID, with closer-to-ground-truth transition dynamics.
Combining AdaIN with Scan yields the highest retrieval scores, lowest FID, and smallest boundary gaps.
Relative to AdaIN alone, adding Scan reduces FID by 6.8\% and $\Delta$PJ and $\Delta\mathrm{AUJ}$ by 73.3\% and 32.7\%, respectively.
Scan alone achieves a smaller diversity gap but weaker fidelity and continuity.
These results show complementary conditioning: AdaIN modulates feature channels, while condition-guided scanning regulates state updates, improving semantic control and transition coherence.\looseness=-2\par
\endgroup

\noindent\textbf{Timestep Sampling Strategy.}
Timestep sampling is an important design choice in generative models.
We compare Uniform sampling with 4 Beta alternatives emphasizing intermediate timesteps $\mathrm{Beta}(2.0,2.0)$, endpoints $\mathrm{Beta}(0.5,0.5)$, high-noise timesteps $\mathrm{Beta}(2.0,1.0)$, and low-noise timesteps $\mathrm{Beta}(1.0,2.0)$.
Each distribution is used for both training time-pair sampling and 4-step inference.
Table~\ref{tab:time_sampling} shows that Uniform achieves the overall best results across reported metrics, supporting its use as our default.
This advantage likely stems from balancing structural recovery at high noise with fine temporal refinement at low noise, addressing the coupled spatial and temporal demands of 3D motion generation.\looseness=-1

\subsection{Motion Editing}
Motion editing enables fine-grained control in interactive 3D motion generation when text or audio alone cannot specify local motion details. Users can demonstrate desired movements, reconstruct them through human mesh recovery~\cite{kocabas2020vibe}, and use the recovered motion as constraints for completion or refinement.
\noindent\textbf{(1) Qualitative Analysis.}
We showcase music-to-dance, where user-performed dance fragments provide intuitive local constraints. As shown in Fig.~\ref{fig:motionedition}, our Adaptive Soft mask generates smooth in-between motions from specified start and end clips. Hard masking still produces reasonable motions within the gap but introduces abrupt boundary discontinuities that disrupt dance fluency.
\noindent\textbf{(2) Quantitative Analysis.}
We evaluate 30 FineDance windows from 19 test recordings using 3 matched seeds, 120-frame gaps, and 4-step sampling.
\textit{Seam} measures mean boundary velocity innovation, and \textit{Jerk} measures mean interior 3rd-derivative magnitude (both lower is better); \textit{Activity} is the generated-to-GT ratio of root-relative joint RMS speed (target 1).
We compare naive binary Hard~\cite{tevet2023human,tseng2023edge}, constant-weight Fixed Soft~\cite{shafir2024prior} ($w=0.5$), and our Adaptive Soft.
Table~\ref{tab:editing_masks} shows that Adaptive Soft achieves the lowest mean Seam and Jerk, reducing them by 84.9\% and 4.2\% relative to Hard while retaining comparable activity, demonstrating improved motion fluidity while maintaining fidelity.
These results demonstrate \textbf{FlashMotion}'s superior boundary continuity over competing methods, making it better suited to a broader range of applications, including windowed streaming generation and human-in-the-loop motion authoring.\looseness=-2
\subsection{Downstream Applications}
\label{sec:applications}

With lightweight post-processing, \textbf{FlashMotion} supports robot motion generation, XR character animation, and mobile motion interaction. We showcase text-to-motion as a representative task (Fig.~\ref{fig:applications}).
\noindent\textbf{(1) Robot Motion Generation.}
Using a GPT-6 Astra-assisted implementation, we retarget generated human motion to a G1 humanoid and track it with a pretrained controller in MuJoCo. The boxing-to-kick example demonstrates the transfer of language-conditioned, coordinated whole-body motion to physical simulation. This workflow can reduce manual trajectory authoring and facilitate robot behavior prototyping.
\noindent\textbf{(2) XR Character Animation.}
We map generated motion to virtual characters through skeletal retargeting. The zombie-walking example demonstrates semantic control of character behavior in an immersive XR environment. As character animation is central to embodied interaction, this workflow can lower motion-authoring costs and support scalable interactive content creation.
\noindent\textbf{(3) Mobile Motion Interaction.}
We convert spoken requests into text conditions and retarget the generated motion to a mobile avatar. With cached text embeddings, quantization, and optimized operators, \textbf{FlashMotion} generates 200 motion frames using 4-step sampling in approximately 400\,ms on iPhone 17, enabling responsive interaction while reducing reliance on cloud inference.

\section{Conclusion}
In conclusion, we present \textbf{FlashMotion}, a co-designed framework for high-quality long-horizon 3D motion generation under a strict inference budget. For generative strategy, \textbf{FlashMotion} proposes a 3D motion-space MeanFlow formulation with kinematic and dynamic consistency constraints, and an Adaptive Soft masking mechanism for temporally coherent motion editing. For model architecture, \textbf{FlashMotion} designs a lightweight non-autoregressive BiMamba architecture with condition-guided state selection and AdaIN-based condition fusion, enhancing cross-modal alignment and generalization to longer motion sequences. Experiments demonstrate that \textbf{FlashMotion} achieves overall state-of-the-art (SOTA) performance in both quality and efficiency across text-to-motion, speech-to-gesture, and music-to-dance tasks.

Potential directions for future research include:
(1) \textbf{Fine-Grained Motion Generation.} Modeling facial expressions and finer finger articulation could enable more detailed and expressive human animation.
(2) \textbf{Robust Generation at Scale.} Reducing motion artifacts in larger-scale training and longer-sequence generation could improve reliability for production use.
(3) \textbf{Efficient Real-World Deployment.} Hardware-aware optimization and end-to-end pipeline integration could facilitate efficient deployment and broader adoption in robotics, XR, and mobile interaction.

\raggedbottom
\IEEEtriggeratref{0}
\bibliographystyle{IEEEtran}
\bibliography{main_arxiv}
\end{document}